\documentclass[10pt,journal,compsoc]{IEEEtran}
\usepackage[noend]{algpseudocode}

\usepackage{algorithmicx,algorithm}

\usepackage{graphicx} 
\usepackage{enumerate}
\usepackage{booktabs}
\usepackage{multirow}
\usepackage{pfnote}
\usepackage{amssymb}
\usepackage{bm}
\usepackage{lineno}

\usepackage{hyperref}
\hypersetup{
	colorlinks=true,
	linkcolor=black,
	filecolor=black,      
	urlcolor=black,
	citecolor=black,
}
\ifCLASSOPTIONcompsoc
  \usepackage[nocompress]{cite}
\else
  \usepackage{cite}
\fi

\ifCLASSINFOpdf
\else
\fi
\usepackage{amsmath}
\begin{document}
\bibliographystyle{IEEEtran}

%
\title{MHNF: Multi-hop Heterogeneous Neighborhood information Fusion graph representation learning}
%
%
%
%

\author{Yundong Sun,Dongjie Zhu*,Haiwen Du and Zhaoshuo Tian,~\IEEEmembership{Member,~IEEE}
\IEEEcompsocitemizethanks{

\IEEEcompsocthanksitem Y. Sun, H. Du and Z. Tian are with the School of Astronautics, Harbin Institute of Technology, Harbin, China, 150001.
E-mail: \{hitffmy@163.com, duhaiwen@126.com, tianzhaoshuo@126.com 
\IEEEcompsocthanksitem D. Zhu is with the School of Computer Science and Technology, Harbin Institute of Technology, Weihai, China, 264209.
E-mail: zhudongjie@hit.edu.cn
\}
}
\thanks{(Corresponding author: Dongjie Zhu, zhudongjie@hit.edu.cn)}}

%
%

\markboth{IEEE TRANSACTIONS ON KNOWLEDGE AND DATA ENGINEERING}%
{Shell \MakeLowercase{\textit{et al.}}: Bare Demo of IEEEtran.cls for Computer Society Journals}
%



\IEEEtitleabstractindextext{%
\begin{abstract}
The attention mechanism enables graph neural networks (GNNs) to learn the attention weights between the target node and its one-hop neighbors, thereby improving the performance further. However, most existing GNNs are oriented toward homogeneous graphs, and in which each layer can only aggregate the information of one-hop neighbors. Stacking multilayer networks introduces considerable noise and easily leads to over smoothing. We propose here a multihop heterogeneous neighborhood information fusion graph representation learning method (MHNF). Specifically, we propose a hybrid metapath autonomous extraction model to efficiently extract multihop hybrid neighbors. Then, we formulate a hop-level heterogeneous information aggregation model, which selectively aggregates different-hop neighborhood information within the same hybrid metapath. Finally, a hierarchical semantic attention fusion model (HSAF) is constructed, which can efficiently integrate different-hop and different-path neighborhood information. In this fashion, this paper solves the problem of aggregating multihop neighborhood information and learning hybrid metapaths for target tasks. This mitigates the limitation of manually specifying metapaths. In addition, HSAF can extract the internal node information of the metapaths and better integrate the semantic information present at different levels. Experimental results on real datasets show that MHNF achieves the best or competitive performance against state-of-the-art baselines with only a fraction of 1/10 $\sim$ 1/100 parameters and computational budgets. Our code is publicly available at \url{https://github.com/PHD-lanyu/MHNF}.

\end{abstract}

}

\maketitle

\IEEEdisplaynontitleabstractindextext

%
\IEEEpeerreviewmaketitle

\IEEEraisesectionheading{\section{Introduction}\label{sec:introduction}}


%
%
%
%
\IEEEPARstart{T}{he} purpose of graph representation learning is to transform the nodes in the graph into low-dimensional vector representations by certain methods. This low-dimensional vector representation can maintain the node features, structural characteristics, and semantic information of the original graph.  Consequently, it can be used for downstream semantic computing tasks, such as node classification, node clustering, and link prediction \cite{cui2018survey}. This can lead to applications  in knowledge question answering systems \cite{jiang2020reasoning}, recommendation systems \cite{wang2019multi}, social influence analyses\cite{qiu2018deepinf}, and other fields. In the early days, the traditional graph representation learning methods (e.g., DeepWalk\cite{perozzi2014deepwalk} and Node2Vec\cite{grover2016node2vec}) used different optimization strategies to optimize the hidden matrix of the neural networks and used this matrix as the low-dimensional vector representation of the network nodes. This type of method needs to maintain a matrix in which the number of rows is equal to the number of nodes and the number of columns is equal to the dimension of the node vectors. Worse still, this type of method does not have the ability of inductive learning, that is, only the nodes that appear in the training set can be represented; they cannot represent new nodes that do not appear in the training set\cite{hamilton2017inductive}.

Later, the emergence of graph neural networks (GNNs)\cite {hamilton2017inductive, kipf2016semi, defferrard2016convolutional, atwood2016diffusion} provided a novel paradigm for graph representation learning and graph data mining tasks. GNNs can be regarded as an extension of the convolutional neural networks (CNNs)\cite{krizhevsky2012imagenet} model in non-Euclidean spaces. This allows non-Euclidean spatial data to be used as the input of the neural networks directly. GNNs use spectral methods\cite {kipf2016semi, defferrard2016convolutional} or spatial domain methods\cite {hamilton2017inductive, atwood2016diffusion} to aggregate information from the one-hop neighbors of the target node, This substantially improves the structure mining ability and local information aggregation ability of graph data. In particular, the introduction of the attention mechanism enables the model to learn the importance of one-hop neighbors relative to the target node according to specific tasks\cite{velivckovic2017graph}. It removes noisy neighborhood information. Therefore, it can bring great improvements to the performance of various downstream tasks.

\begin{figure}[t]
  \centering
  \includegraphics[width=3.5in,trim=0 40 0 20]{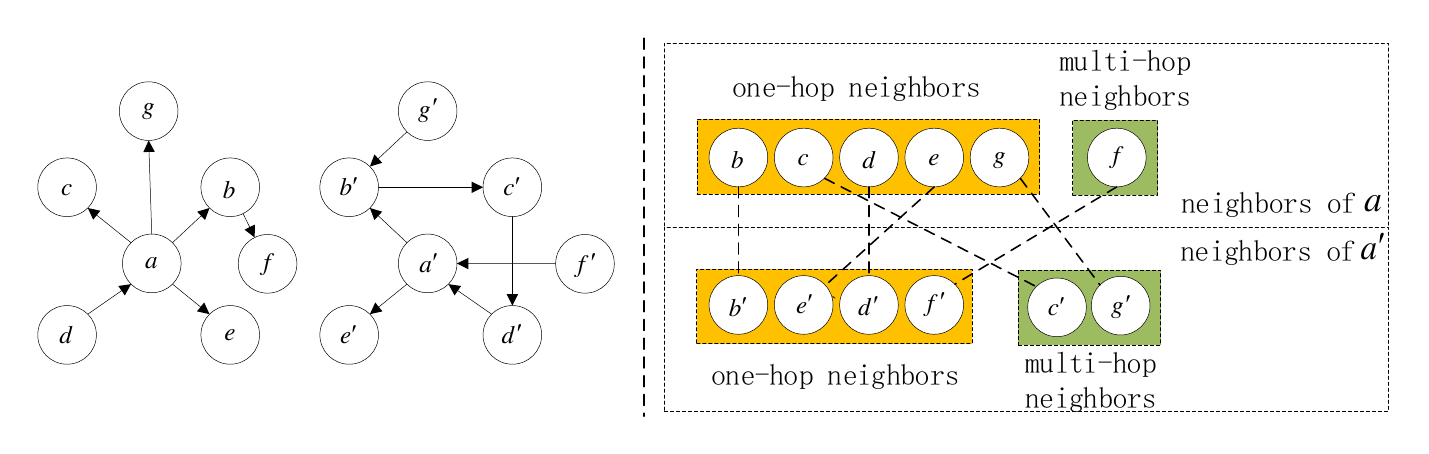}
  \caption{Schematic diagram of the non-isomorphism analysis of one-hop neighbors.}
  \label{figure1}
  \end{figure}

Although the attention mechanism improves the performance of the GNNs, the single-layer network can only aggregate information of one-hop neighbors of the target node. However,  the multihop neighbors' information also plays an important role in representing the target node. Pujara et al. \cite{pujara2013knowledge} found that due to the incompleteness of data and the diversity of patterns, the corresponding entity usually has different neighborhood structures in different linguistic knowledge graphs. As shown in Fig. 1, the one-hop neighbors of a and a' are different. Sun et al. \cite{sun2020knowledge} analyzed and summarized DBpedia, an open-source knowledge graph dataset, and found that among the corresponding nodes under different language knowledge graphs, the proportion of nodes with different one-hop neighbors reached approximately 90\%. This non-isomorphism can easily lead to excessive differences in the representation of corresponding entities generated by GNNs. This will seriously affect the performance of subsequent tasks such as entity alignment. The addition of multihop neighbors will be able to compensate for this representation mismatch problem (see the right of Fig. 1) \cite{sun2020knowledge}.

\begin{figure}[t]
  \centering
  \includegraphics[width=3.5in,trim=0 40 0 20]{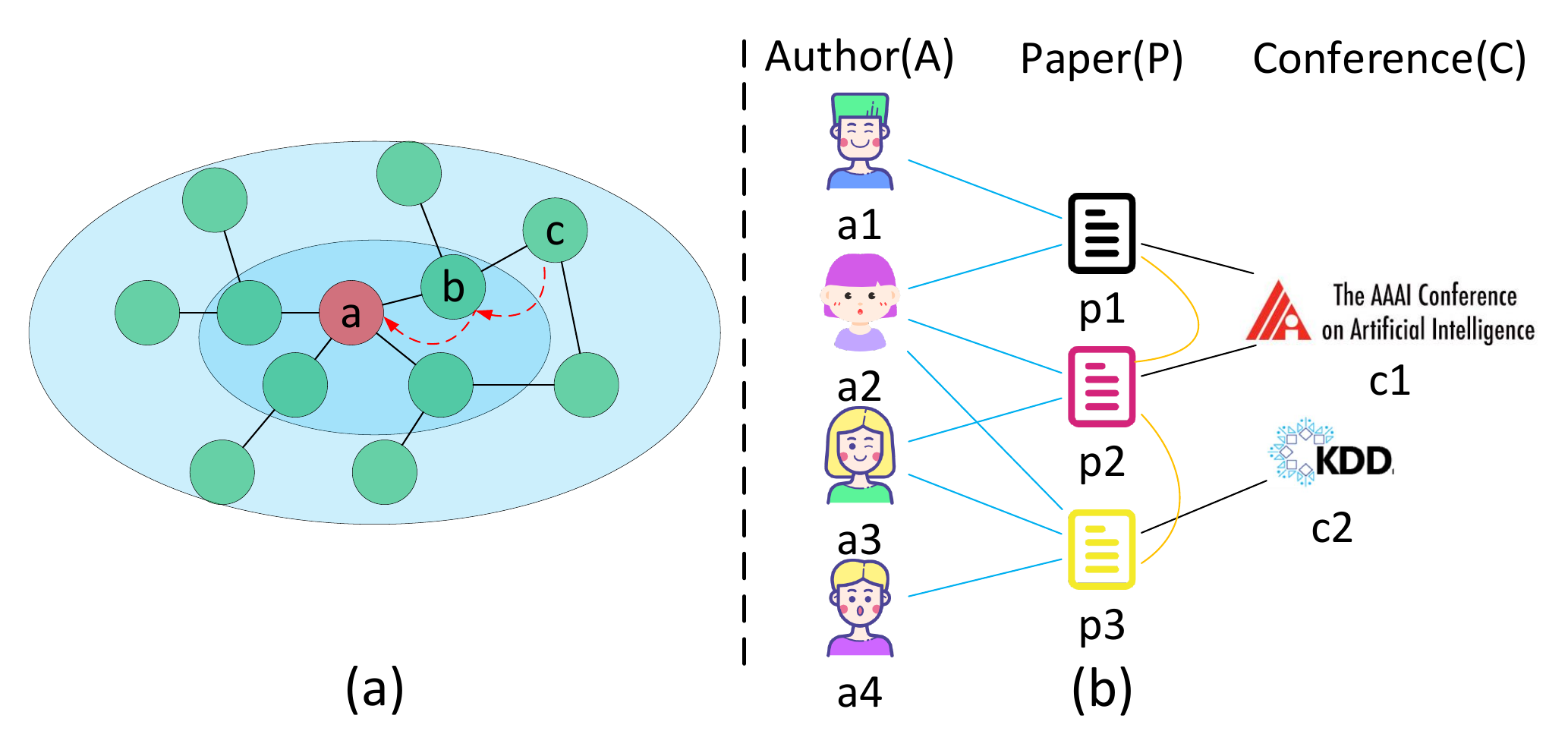}
  \caption{Schematic diagram of multihop neighbors and the heterogeneous relationship between nodes. (a) The solid lines denote the direct information aggregation of one-hop neighbors, and the red dashed lines denote the aggregation of multihop neighborhood information in a transitional manner by stacking existing multilayer GNNs. (b) A, P, C denote different node types, and different colored edges denote different relationships.}
  \label{figure2}
  \vspace{-0.5cm}
  \end{figure}

At the same time, the above methods are only applicable to homogeneous networks, in which the nodes and edges are regarded as a single type. However, most graph data in real scenarios are heterogeneous (such as the academic network shown in Fig.\ref{figure2}(b)). There will be multiple types of nodes (scholars, papers, journals, and conferences) and edges (the signature  relationship between scholars and papers, the citation relationship between papers, and the publication relationship between papers and journals) \cite {dong2017metapath2vec, shi2018heterogeneous}. The different types of nodes and complex interactions enable heterogeneous graphs to contain more information than homogeneous graphs. Ignoring this heterogeneous information will make it difficult to mine the characteristics of different types of nodes and the complex semantic relationships between them. This will inevitably degrade  the performance of downstream tasks. In this context, Metapath2vec\cite{dong2017metapath2vec} designed a random walk strategy based on metapath\cite{sun2012mining} to extract nodes with different relationships, and then the skip-gram model\cite{mikolov2013efficient} was used for graph embedding. Semantic-level attention was introduced in \cite{velivckovic2017graph} to extend the GAT model to fuse the representation results under different metapaths so that HAN\cite{wang2019heterogeneous} can better integrate richer semantic information. However, the existing metapath-based methods need to manually specify the metapath, and it is impossible to autonomously extract valuable metapaths according to concrete tasks.

Based on the above analysis, GNNs currently show promising performance in various tasks in the field of graph data mining. However, when dealing with  heterogeneous data, state-of-the-art methods have not solved adequately the problems involving such type of data. Further, these methods may not even be aware of the following challenges, which we address in this paper:

\begin{itemize}
	\item \textbf{(C1) How can multihop neighbors' information be efficiently aggregated?} A single-layer neural network can only aggregate information of one-hop neighbors (from b to a in Fig. 2(a)). More recently, some state-of-the-art approaches\cite{xia2021knowledge,qin2021relation,jing2021hdmi,jin2020multi} have focused on capturing various relationships among nodes during the message passing process. However, these methods cannot perform well in effectively aggregating the information of multihop neighbors in heterogeneous graphs. The aggregation of multihop neighborhood information can only be achieved by stacking multilayers of GNNs (in a transitional manner) in these methods\cite{zhang2019heterogeneous}. As shown in Fig. 2(a), the information of node c is first transmitted to node b and then to node a. However, previous studies have shown that although the stacking of multilayer GNNs can aggregate multihop neighbor nodes in a transitional manner, simply increasing layers of the networks will bring about problems such as the vanishing gradient and oversmoothing \cite{li2019deepgcns}.

	\item \textbf{(C2) How can multihop neighbors be autonomously extracted under heterogeneous relations?} Although a few researchers\cite{xue2020multi} have devoted attention to the information aggregation of multihop neighbors, they are rooted in the field of homogeneous graph data. Most of the existing heterogeneous graph representation methods\cite{dong2017metapath2vec, wang2019heterogeneous} need to manually specify the metapath based on experience or domain knowledge in advance  (for example, the metapath of APCPA in Fig. 2(b)). They cannot learn the corresponding hybrid metapath autonomously according to the target task. Therefore, the performance of the downstream task is very sensitive to the specified metapath. Admittedly, GTN\cite{yun2019graph} can learn soft composite relations for generating useful multihop metapaths while automatically controlling their lengths. However, it cannot efficiently integrate different-hop and different-path neighborhood semantic information. This limitation will be verified in our experiments (Sections 5.4 and 5.5). 
	
	\item \textbf{(C3) How can multilevel information of neighbors be comprehensively integrated?} At present, most metapath-based heterogeneous graph representation methods\cite {dong2017metapath2vec, wang2019heterogeneous, hu2020heterogeneous} only focus on the node information at the two ends of the metapath and ignore the information of the intermediate ones in the path. However,  such information is also important for representing learning\cite{fu2020magnn}. As shown in Fig. 2(b), two authors in the metapath of APA may cooperate on different papers (a3, a2). Ignoring the information concerning the papers will not be conducive to mining the precise relationship between the two authors. In addition, there is a lack of effective integration of different metapaths on this basis. This is however needed since different metapaths represent different semantic information. For example, APA in Fig.\ref{figure2}(b) represents two authors jointly publishing an article, and APCPA represents two authors whose articles were published at the same conference or journal. Hence the efficient integration of the above semantics at different levels is essential for the improvement of task performance. In this light, MAGNN\cite{fu2020magnn} can aggregate the intra-metapath information and inter-metapath information successively. However, it ignores the differences between the neighbors of different hops and the correlation between the same hops. Most importantly, the metapath of MAGNN still needs to be specified manually and cannot be learned based on the target task.
	
\end{itemize}

To tackle the above challenges, this paper starts with the problem of multihop neighborhood information aggregation in heterogeneous graphs. It then proposes a representation learning method of multihop heterogeneous neighborhood information fusion networks (MHNF). Specifically, we first build a Hybrid Metapath Autonomous Extraction (HMAE) model, which can autonomously learn valuable hybrid metapaths using which one can efficiently extract multihop neighbors with different relationships. Based on the learned hybrid metapath, we propose a Hop-Level Heterogeneous Information Aggregation (HLHIA) model, which selectively aggregates the information of different-hop neighbors within the same hybrid metapath. Finally, a Hierarchical Semantic Attention Fusion (HSAF) model is constructed, which includes an information fusion model for different hops within one hybrid metapath and an information fusion model for different hybrid metapaths, which can efficiently integrate the information of the two levels. The contributions of this article can be summarized as follows:

\begin{itemize}
  \item [1)] This paper proposes the novel concept of a hybrid metapath for the first time. Different from the existing metapath notion, the hybrid metapath is a soft combination of multiple single relations learned for the concrete task. Compared with the existing metapath idea, it contains potentially richer and more valuable interactions for the target task.
      
  \item [2)] Based on the proposed hybrid metapath, the HMAE model proposed in this paper can autonomously learn valuable hybrid metapaths of specified lengths to realize the efficient extraction of multihop neighbor nodes with different distances. It can solve the problem of information aggregation from multihop neighbors for GNNs and can autonomously learn hybrid metapaths of different lengths according to the task thus eliminating the limitation of manually specified metapaths.
  
  \item [3)] Based on the HLHIA model proposed in this paper, the information contained in different hops in the same hybrid metapath is selectively aggregated, and the aggregated information of different hops in each hybrid metapath is obtained. Furthermore, the model proposed in this paper can fully determine the importance of different hops and different metapaths. This solves the problem that the existing methods cannot, namely,  extract the information of the internal nodes of the metapath. At the same time, it can better understand and fuse semantic information at different levels.
 
  \item [4)] A large number of experiments have been conducted on three real heterogeneous graph datasets. The results show that the method proposed in this paper is superior to the state-of-the-art methods in node classification and node clustering tasks while using only 1/10 $\sim$ 1/100 number of parameters and computational budget. At the same time, outstanding performance has also been achieved in attention analysis experiments, visualization experiments, and parameter stability experiments.

\end{itemize}

The rest of this article is organized as follows: We start with some preliminaries in Chapter 2. Chapter 3  discusses related researches about heterogeneous graph neural networks. Chapter 4 introduces our proposed method in detail. We then conduct extensive experiments, compare with the benchmark methods, and discuss and analyze the results in Chapter 5. Finally, we conclude in the Conclusion section.

\section{Releted Work}\label{sec:releted work}

In this section, we will discuss and analyze related researches about heterogeneous graph neural networks.

Some early GNNs\cite {bruna2013spectral, shuman2013emerging, kipf2016semi, hamilton2017inductive,velivckovic2017graph, schlichtkrull2018modeling, zhou2019multi, geng2019spatiotemporal} are only applicable to homogeneous graphs while most data in real scenarios are difficult to characterize by homogeneous graphs\cite{cen2019representation}. For example, scholar nodes and paper nodes in an academic graph cannot be treated homogeneously. Similarly, the relationship between scholars and papers and the publication relationship between papers and journals should also be treated differently. Indeed, a number of attempts have been made to improve the representation ability of GNNs for heterogeneous graphs. The HAN model\cite{wang2019heterogeneous} is an optimized version based on the GAT model\cite{velivckovic2017graph}. The model first specifies multiple different metapaths, leverages multiple GATs to learn the node representation of the homogeneous network generated by each path, and then designs semantic-level attention to merge the representation results under different metapaths. The experimental results demonstrate that the model can better integrate richer semantic information. 

Recently, some approaches\cite{xia2021knowledge,qin2021relation,jing2021hdmi,jin2020multi} that are also focusing on capturing relationships among nodes during the message passing process have been proposed. These works differentiate graph connection relationships and design relation-aware message passing methods (eg: self-attention mechanism\cite{qin2021relation}, learning node embedding on multiplex networks in a self-supervised way\cite{jing2021hdmi}) for graph representation learning with heterogeneous relational structures. These methods can well mine and leverage the complex and diverse relationship information between nodes, thereby improving the performance of downstream tasks. 
However, these methods can only aggregate the information of multihop neighbors by stacking multilayer GNNs. Some existing studies have shown that simply stacking multiple layers in GNNs will cause problems such as vanishing gradients and oversmoothing of node attributes\cite {li2019deepgcns, xue2020multi}. To tackle this problem, G. Li et al.\cite{li2019deepgcns} borrow the idea of ResNet\cite{he2016deep} in image processing using which GNNs can also build dozens of layers of deep network structures as in CNNs\cite {he2016deep, huang2017densely}. However, such a transitional way of aggregating information will cause information loss or the introduction of noise. Xue et al.\cite{xue2020multi} aggregate neighborhood information at different hops by an attention mechanism and can selectively extract significant multihop features. However, it can only be applied to homogeneous graphs. Hence there is a need to study the problem of effective information aggregation of multihop neighbors in heterogeneous graphs. For more literatures on heterogeneous graph neural networks, please refer to the reviews \cite{yang2020heterogeneous} and \cite{dong2020heterogeneous}.
The method of MHNF in this paper not only pays attention to the relationship between nodes during information passaging but also can efficiently mine and effectively aggregate the information of multihop neighbors in heterogeneous graphs. This is also the main advantage of MHNF compared to the above research works.

\begin{figure*}[t]
  \centering
  \includegraphics[width=5in,trim=0 40 0 20]{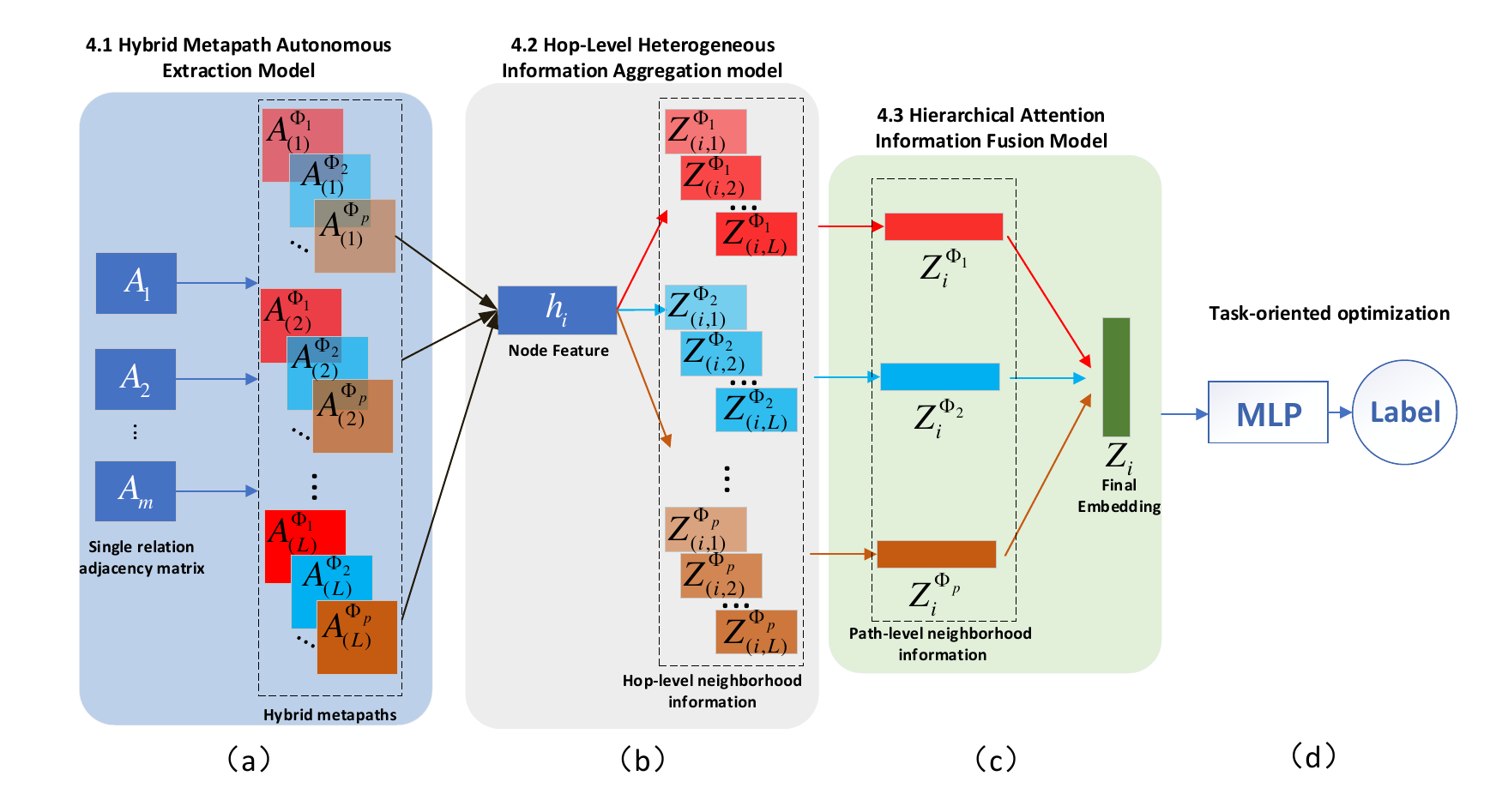}
  \caption{Schematic diagram of the overall framework of the proposed MHNF model. (a) the process of hybrid metapath extraction. We input the adjacency matrix of every single relationship and use the hybrid metapath autonomous extraction model to extract the adjacency matrix of different hops under different hybrid metapaths. Different colors represent different hybrid metapaths, and the color depth represents the number of hops. (b) the process of hop-level information aggregation. The hop-level heterogeneous information aggregation model is used to aggregate information of different hop neighbors within the same hybrid metapath. (c) the process of hierarchical semantic attention information fusion. Using the hierarchical semantic attention fusion model, the information of different hops in the path is first fused to obtain path-level neighborhood information. Then the information of different paths is fused to obtain the final representation. (d) the process of learning and optimizing for specific tasks.}
  \label{figure3}
  \vspace{-0.4cm}
  \end{figure*}
  
\section{Preliminaries}\label{sec:Preliminaries}

In this section, we briefly describe the concepts, definitions and related variables used in this paper.

\textit{Definition3.1} \textbf{Heterogeneous Graph.} The difference between heterogeneous and homogeneous graphs is that heterogeneous graphs contain either multiple types of nodes or multiple types of edges.  In contrast, there is only one type of node and edge in homogeneous graphs. If the heterogeneous graph is formulated as $G = (V,E)$, with a node type mapping function ${f_v}:V\to {T}$ and an edge type mapping function ${f_e}:E \to R$, then $|T| + |R| > 2$, where $V$ and $E$ represent the set of nodes and edges, $T$ and $R$ represent the set of node types and edge types, respectively. For example, the citation network in Fig.\ref{figure2}(b) is a heterogeneous graph, which includes three types of nodes: Author, Paper, and Conference, and multiple types of edges, such as Author-Paper, Paper-Paper, and Paper-Conference. Compared with homogeneous graphs, heterogeneous graphs are more in line with applications in real scenarios and have stronger representation capabilities.

\textit{Definition3.2} \textbf{Metapath.} Metapath is a path with a special semantic relationship in heterogeneous graphs, usually formulated in the form of ${T_1}\mathop  \to \limits^{{R_1}} {T_2}\mathop  \to \limits^{{R_2}} ...\mathop  \to \limits^{{R_l}} {T_{l + 1}}$ and abbreviated as ${T_1}{T_2}...{T_{l + 1}}$. This path indicates that nodes ${1}$ and ${{l + 1}}$ have a composite relationship $R = {R_1} \circ {R_1} \circ ...{R_l}$, where $T_{i}$ represents the type of node $i$ and $R_{i}$ represents the type of edge $i$. For example, path a1p1c1p2a3 in Fig. 2(b) indicates that there is a composite relationship $R=APCPA$ between a1 and a3. In a real scenario, the papers a1 and a3 were published at the same conference. However, this kind of composite relationship cannot be discovered in homogeneous graphs.

\textit{Definition3.3} \textbf{Hybrid metapath.} The metapath in \textit{Definition3.2} must be manually specified based on domain knowledge. We propose a novel concept called Hybrid metapath. It is no longer a human-designated path but rather, a soft combination of various relationships identified at each step by the model according to the specific task. In this case, the combination relationship $R$ (see Definition 3.2) becomes $R' = {R_1}^\prime  \circ {R_2}^\prime  \circ ... \circ {R_l}^\prime  = (\sum\limits_{{m_1} \in T} {{w_{{m_1}}}} {R^{{m_1}}}) \circ (\sum\limits_{{m_2} \in T} {{w_{{m_2}}}} {R^{{m_2}}}) \circ ... \circ (\sum\limits_{{m_l} \in T} {{w_{{m_l}}}} {R^{{m_l}}})$. Since the hybrid relationship ${T_1}\mathop  \to \limits^{{R_1}^\prime } {T_2}\mathop  \to \limits^{{R_2}^\prime } ...\mathop  \to \limits^{{R_l}^\prime } {T_{l + 1}}$ is learned for specific tasks, it can mine richer and more valuable relationships for downstream tasks.

\textit{Definition3.4} \textbf{Metapath based Neighbors.} In a heterogeneous graph, given a node $i$ and a metapath $\Phi$, starting from the node $i$, the nodes $j$ that can be reached along the path $\Phi$ form a set $N_i^\Phi$ called the neighbors of the node   under metapath  $\Phi$.

\textit{Definition3.5} \textbf{Metapath based Multi-Hop Neighbors.} In a heterogeneous graph, given a node $i$ and a metapath $\Phi$, starting from the node $i$, the nodes $j$ that can be reached across $l$ edges along the path $\Phi$ form a set $N_{i,l}^\Phi$. $N_{i,l}^\Phi$ are the $l$-hops neighbors of the node $i$ under metapath $\Phi$.

\textit{Definition3.6} \textbf{Inner-Metapath Nodes.} The node set $\{{2},{3},...,{l}\}$ except for the starting node ${1}$ and the destination node ${{l + 1}}$ are the inner nodes of the metapath ${T_1}\mathop  \to \limits^{{R_1}} {T_2}\mathop  \to \limits^{{R_2}} ...\mathop  \to \limits^{{R_l}} {T_{l + 1}}$.

The notations used in the article are summarized in Table 1.
\vspace{-0.5cm}
\begin{table}[h]
\setlength{\abovecaptionskip}{-0.2cm}
\caption{Notations and Explanations.}

\label{tab1}
\center
\begin{tabular}{p{30pt}p{180pt}}
\hline
Notation & Explanation                                                                               \\ \hline
$\Phi $        					& Hybrid metapath. \\
$T_{i}$								& The type of node $i$.\\
$h$        						& The initial feature matrix of the node, each row ${h_i} \in {R^d}$ of matrix $h$ denotes the representation of a $d$-dimensional feature vector of node $i$.                                              \\
$h'$        						& Projected node feature.                                                         \\
${A_m}$        					& Adjacency matrix under relation $m$.                                                      \\
$A_{(l)}^\Phi$        			& Hybrid metapath based $l$-hop adjacency.                                                                            \\
${w_m}$       					& Importance of relation $m$.                                            \\
$\beta _{(i,l)}^\Phi$        & Hybrid metapath based hop-level attention for $l$-hop neighbors of node $i$.                                                                   \\
${\beta _\Phi }$        		& Weight of Hybrid metapath $\Phi $.                                                    \\
$N_i^\Phi $        				& Hybrid Metapath based neighbors of node $i$.                                                                   \\
$N_{i,l}^\Phi $        		& Hybrid Metapath based $l$-hop neighbors of node $i$.                                                         \\
$Z_{(i,l)}^\Phi$        		& Semantic embedding of $N_{i,l}^\Phi$                                                                 \\
$Z_i^\Phi$        				& The comprehensive embedding of node $i$ for $\Phi $.                                                                  \\
${Z_i}$        					& The final embedding of node $i$.                                                                       \\
$P$        						& The hybrid metapath set                    \\ \hline                                                                    
\end{tabular}
\vspace{-1cm}
\end{table}

\section{The proposed method.}\label{sec:The proposed method.}

In this section, we will introduce and analyze the proposed MHNF model in detail. To extract and fuse information at different semantic levels efficiently, this paper proposes a hierarchical model structure, which fuses the hierarchical neighborhood information of different hops within a hybrid metapath and the semantic information of different hybrid metapaths. The overall model framework is shown in Fig. \ref{figure3}. First, we construct a hybrid metapath autonomous extraction (HMAE) model (Section 4.1), which can autonomously learn valuable hybrid metapaths of a specified length according to the given task. This model achieves efficient extraction of multihop neighbors of different relations, as shown in Fig. \ref{figure3} (a). Based on the learned hybrid metapath, a Hop-Level Heterogeneous Information Aggregation (HLHIA) model within a single path is defined (Section 4.2), which can attentionally aggregate the information of different-hop neighbors within the same hybrid metapath, as shown in Fig. \ref{figure3} (b). Furthermore, we propose a hierarchical semantic attention fusion (HSAF) model (Section 4.3). This model can fuse the neighborhood information of different hops in a single path and the semantic information of different paths, as shown in Fig. \ref{figure3} (c). Finally, we feed the final embeddings into the multilayer perceptron (MLP) network to tune the model parameters using node classification tasks, as shown in Fig. \ref{figure3}(d). In the following subsections, we will describe each substructure in detail.

\subsection{HMAE: Hybrid Metapath Autonomous Extraction model.}

Most of the existing heterogeneous graph representation learning methods are based on metapaths to extract high-order neighbors. However, these methods need to manually specify the metapath. Thus each specified metapath represents a fixed semantic interaction, which cannot be tuned and optimized according to the target task. In real scenarios, the relationship between two nodes is complex and diverse, and the roles and importance of different relationships in different tasks or scenarios are also different. Taking the heterogeneous academic graph in Fig.\ref{figure2}(b) as an example, if we pay more attention to analyzing the cooperative relationship between authors, the path of APA should account for a higher proportion. Similarly, if we pay more attention to analyzing the differences in research fields between authors, the proportion of APCPA should be higher. Therefore, the key is to combine the semantic paths of different relationships in a soft coding manner when constructing a hybrid interaction path. This is also the core insight of this paper.

In this subsection, we propose a hybrid metapath autonomous extraction (HMAE) model. Its goal is to learn the hybrid metapath autonomously for the given task, thereby extracting hybrid heterogeneous neighbors. The overall framework of HMAE is shown in Fig.\ref{figure4}. First, we need to extract different single relationship matrices from the original heterogeneous network, such as the relationship matrices  PA, AP, PC, and CP, as shown in Fig.\ref{figure4} (a). Then, the process of extracting the one-hop hybrid adjacency matrix, as shown in Fig.\ref{figure4} (b), is performed. Specifically, a one-dimensional convolutional neural network is utilized to perform a weighted fusion operation on different single relation matrices:

\begin{figure}[t]
  \centering
  \includegraphics[width=3.5in,trim=0 40 0 40]{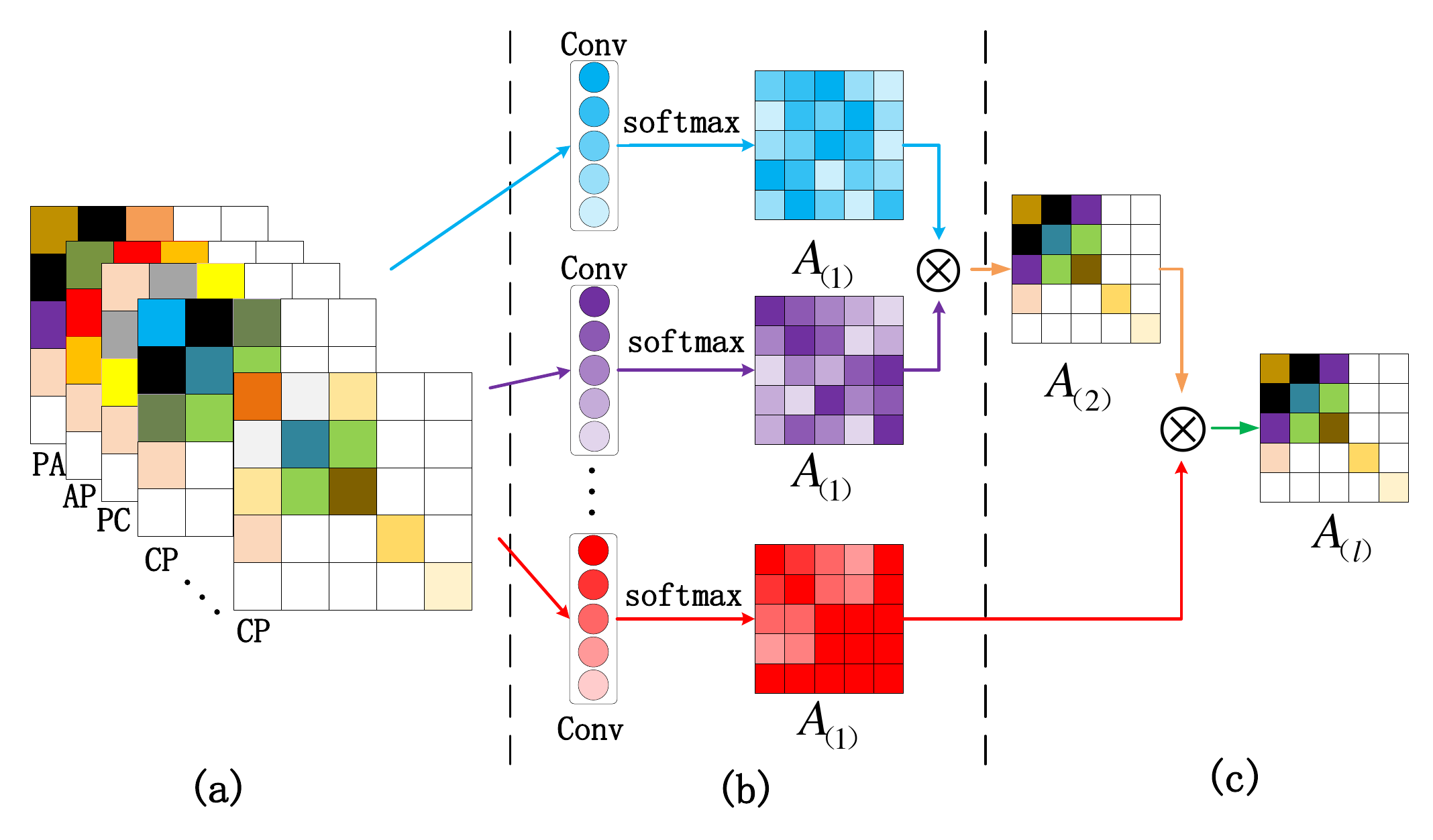}
  \caption{The framework of HMAE. (a) many single relationship matrices of the original heterogeneous graph, where PA and AP denote different relationships. (b) the extraction process of the hybrid one-hop neighbors,$Conv$ denotes the one-dimensional convolutional neural network and${A_{(1)}}$represents the one-hop hybrid adjacency matrix. (c) the extraction process of the multihop hybrid adjacency matrix. ${A_{(2)}}$ and ${A_{(l)}}$ represent the two-hop hybrid adjacency matrix and the $l$-hop hybrid adjacency matrix, respectively. The color depth in the hybrid adjacency matrix denotes the different weights.}
  \label{figure4}
    \vspace{-0.5cm}
  \end{figure}

\begin{equation}\label{eq1}
\begin{aligned}
{A_{(1)}} = F\left( {A;{W_c}} \right) = Conv(A;{W_c})
\end{aligned}
\end{equation}

where $Conv$ represents the convolution operation, ${W_c} \in {\mathbb{R}^{|m| \times 1}}$ is the learnable parameter matrix of the convolutional neural network and $|m|$ represents the number of single relations. Through the above convolution operation, the fusion of different relationships is realized. Furthermore, we perform a softmax operation on each row of the extracted hybrid relationship matrix:

\begin{equation}\label{eq2}
\begin{aligned}
{A_{(1)}}=softmax({A_{(1)}})
\end{aligned}
\end{equation}

After the one-hop hybrid adjacency matrix is obtained, the multilayer networks can realize the extraction of the hybrid element path of the specified length:

\begin{small}
\begin{equation}\label{eq3}
\begin{aligned}
{A_{(l)}}  &= \Pi _{i = 1}^l{A_{(i)}} \\&= (\sum\limits_{{m_1} \in T} {{w_{{m_1}}}{A_{{m_1}}}} )(\sum\limits_{{m_2} \in T} {{w_{{m_2}}}{A_{{m_2}}}} )...(\sum\limits_{{m_l} \in T} {{w_{{m_l}}}{A_{{m_l}}}})
\end{aligned}
\end{equation}
\end{small}

where ${w_m}$ is the fusion weight of relation $m$ learned by $Conv$ and $\Pi$ represents matrix multiplication. Equation (3) shows that each layer of networks is equivalent to performing a weighted summation operation on different relations. The uniqueness of our approach is that the weights are not manually specified. Rather, they are obtained by the model's task-oriented autonomous learning. In this way, by stacking $l$-layer HMAEs, the model can autonomously obtain a hybrid metapath with $l$-hops. In each layer, the corresponding hybrid heterogeneous adjacency matrix is recorded, which is used to aggregate neighborhood information at different hops within the path as described in Section 4.2.
\begin{figure}[h]
  \centering
  \includegraphics[width=3.5in,trim=0 40 0 20]{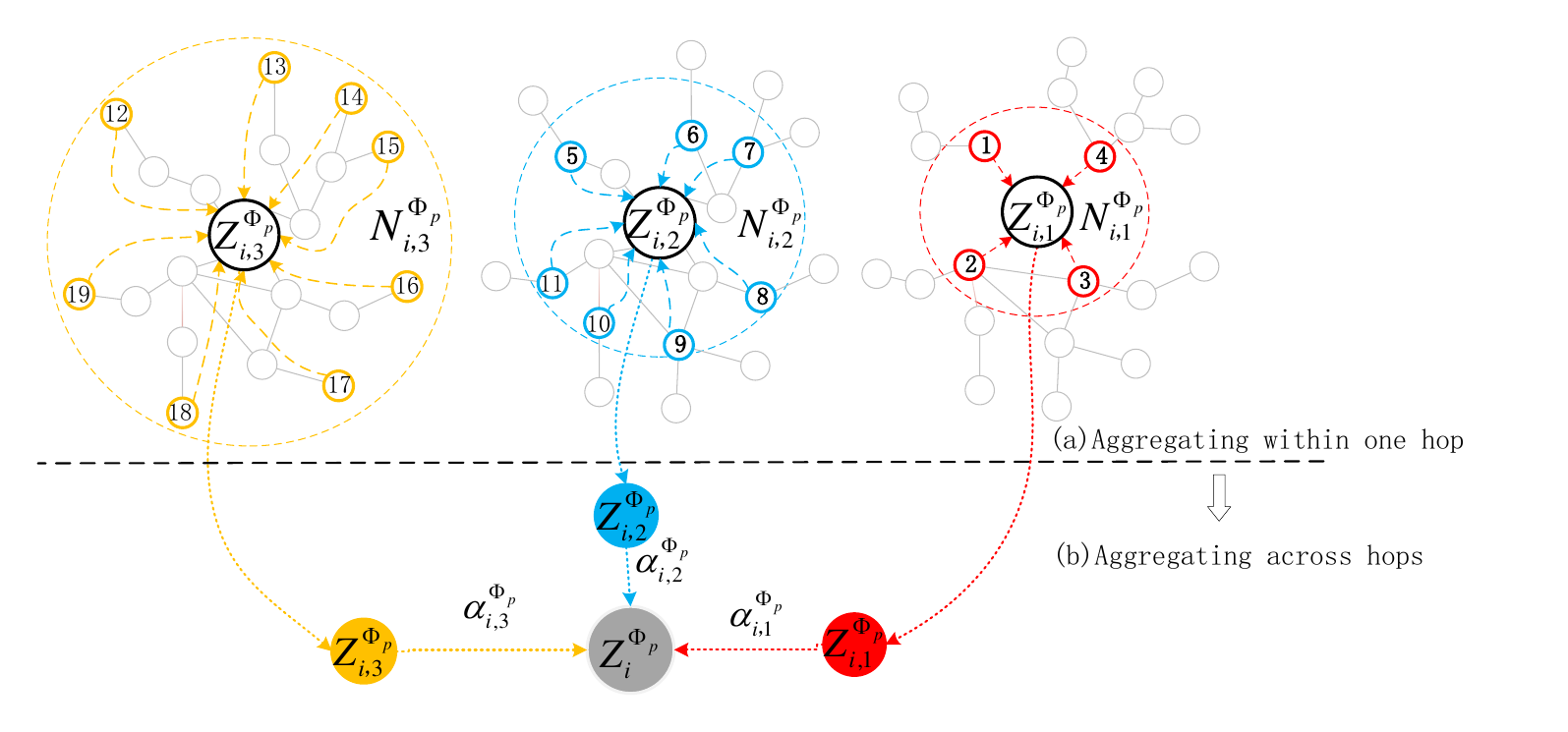}
  \caption{(a) the process of aggregating the information of single-hop neighbors within a hybrid metapath. (b) the information fusion process of different hops neighbors within a hybrid metapath based on the attention mechanism.}
  \label{figure5}
    \vspace{-0.5cm}
  \end{figure}
\subsection{HLHIA: The Hop-Level Heterogeneous Information Aggregation model within one hybrid metapath.}

In Section 4.1, we extracted the hybrid metapath with $l$-hops. In each layer, the corresponding hybrid heterogeneous adjacency matrix ${A_{(i)}}$ is recorded. The existing metapath-based heterogeneous graph representation methods only focus on the node information at the head and tail of the metapath, ignoring the information of the inner-metapath nodes (see \textit{Definition3.6}). Transforming and concatenating all node information in the path is a plausible scheme. However, in real scenarios, different nodes will have different importance to the representation of the path. Hence such a scheme cannot effectively distinguish the importance of different nodes in the path. It can also easily introduce noise which will degrade the accuracy of node information representation.

To solve this problem, this paper proposes a strategy of hop-level information aggregation within one hybrid metapath, as shown in Fig.\ref{figure5}. Based on the adjacency relationship of each hop in the hybrid metapath obtained in Section 4.1, this section formulates a hop-level heterogeneous information aggregation model within one hybrid metapath to achieve efficient integration of overall nodes in the path. The schematic diagram of the model is shown in Fig.\ref{figure5}(a).

First, in heterogeneous graph data, different types of nodes have different attribute spaces, and information from different spaces needs to be projected to the same space:
\begin{equation}\label{eq4}
\begin{aligned}
{h'_i} = {M_{{\tau _i}}}{h_i}
\end{aligned}
\end{equation}

where ${h_i}$ is the original feature of node $i$, ${h_i}^\prime$ is the projected feature of node $i$, ${M_{{\tau _i}}}$ is the projecting parameter, and ${\tau _i}$ represents the type of node $i$. After mapping the features of all nodes to the same space, we attempt to aggregate the node information of different hops in the path:
\begin{equation}\label{eq5}
\begin{aligned}
Z_l^{{\Phi _p}} = \sigma [{(D_{(l)}^{{\Phi _p}})^{ - 1}}A_{(l)}^{{\Phi _p}}h'{W^{{\Phi _p}}}]
\end{aligned}
\end{equation}

where $Z_l^{{\Phi _p}}$ is the aggregated information from the $l$-hop neighbors under the hybrid metapath ${\Phi _p}$, and the ith row $Z_{i,l}^{{\Phi _p}}$ of $Z_l^{{\Phi _p}}$ represents the aggregated information of node $i$, $A_{(l)}^{{\Phi _p}}$ is the adjacency matrix with $l$-hops under ${\Phi _p}$, $D_{(l)}^{{\Phi _p}}$ is the degree matrix of $A_{(l)}^{{\Phi _p}}$, $h' \in {\mathbb{R}^{N \times d}}$ is the feature matrix of nodes after projection, $N$ is the number of nodes, $d$ is the dimension of the projected feature, ${W^{{\Phi _p}}} \in {\mathbb{R}^{d \times d}}$ is the learnable matrix parameter and $\sigma$ is the activation function of $ReLu$.

\subsection{HSAF: Hierarchical Semantic Attention Fusion model.}

After obtaining the information of different hops under each path in Section 4.2, the information of different hops and different paths needs to be fused. This subsection integrates two levels of information: (1) the fusion of information from different-hop neighbors in one path and (2) the fusion of different-paths information. First, we fuse the information of different hops in one path obtained in Section 4.2. The information of different hops can reflect the different characteristics of nodes. To efficiently distinguish the importance of different hops to the expression of node information, an attention fusion model is designed and implemented. Using this model, different hop-attention weights can be learned, and then the learned weights can be used to selectively fuse different hops of information. Specifically, the following equation is used to learn the importance of different hops:

\begin{equation}\label{eq6}
\begin{aligned}
\alpha _{i,l}^{{\Phi _p}} = \sigma [{\delta ^{{\Phi _p}}}\tanh ({W^{{\Phi _p}}}Z_{i,l}^{{\Phi _p}})]
\end{aligned}
\end{equation}

where $\alpha _{i,l}^{{\Phi _p}}$ is the importance of the information ($Z_{i,l}^{{\Phi _p}}$) of the ${{l}^{th}}$-hop neighbors of the node $i$ under the path ${\Phi _p}$, and ${\delta ^{{\Phi _p}}}$ represents the learnable matrix. Furthermore, the softmax function is used to normalize the importance coefficient of the information of each hop:

\begin{equation}\label{eq7}
\begin{aligned}
\beta _{i,l}^{{\Phi _p}} = \frac{{\exp (\alpha _{i,l}^{{\Phi _p}})}}{{\sum\nolimits_{j = 1}^L {\exp (\alpha _{i,j}^{{\Phi _p}})}}}
\end{aligned}
\end{equation}

Finally, the fusion of the information of different hops and the information representation of node $i$ under the hybrid path ${\Phi _p}$ is obtained:

\begin{equation}\label{eq8}
\begin{aligned}
Z_i^{{\Phi _p}} = \sum\nolimits_{l = 1}^L {\beta _l^{{\Phi _p}}} Z_l^{{\Phi _p}}
\end{aligned}
\end{equation}

Through the above steps, the fusion of the information of each layer in the path is realized. In the next step, a similar scheme is used to integrate the information of different paths. We use Equation (9) to learn the importance of different paths.

\begin{equation}\label{eq9}
\begin{aligned}
{\alpha _{i,{\Phi _p}}} = \sigma [\delta \tanh (WZ_i^{{\Phi _p}})
\end{aligned}
\end{equation}

Similarly, the Softmax function is used to normalize the importance coefficient of each path information:

\begin{equation}\label{eq10}
\begin{aligned}
{\beta _{i,{\Phi _p}}} = \frac{{\exp ({\alpha _{i,{\Phi _p}}})}}{{\sum\nolimits_{p' \in P}^{} {\exp ({\alpha _{{\Phi _{p'}}}})} }}
\end{aligned}
\end{equation}

Finally, the information of different paths is fused to obtain the final information representation of the node:
\begin{equation}\label{eq11}
\begin{aligned}
{Z_i} = \sum\limits_{p \in P} {{\beta _{i,{\Phi _p}}}{Z_{i,{\Phi _p}}}}
\end{aligned}
\end{equation}

where ${Z_i}$ is a comprehensive information representation containing different-hop neighborhood information under different hybrid semantic paths. We can use the comprehensive information of the node to perform different tasks and design the corresponding loss function. This paper designs the node classification cross-entropy loss function for the node classification task:
\begin{equation}\label{eq12}
\begin{aligned}
Loss =  - \sum\limits_{i = 1}^C {{y_i}\log (F({Z_i}))}
\end{aligned}
\end{equation}

where $C$ represents the number of node labels, ${y_i}$ is the probability that the true label of the node is $i$, ${y_i}$ is 1 or 0, and $F({Z_i})$ represents the prediction result of the classification model using the hybrid vector information ${Z_i}$ of node $i$. Based on the above loss function, the gradient descent algorithm is used to optimize the model. The specific algorithmic process is shown in Algorithm 1.
\begin{algorithm}[h]

  \caption{: MHNF} 
  \hspace*{0.02in} {\bf Input:}\\ 
  \hspace*{0.4in}The heterogeneous graph $G=(V,E)$.\\
  \hspace*{0.4in}The node feature $h = \{ {h_i},\forall i \in V\} $.\\
  \hspace*{0.4in}The single relation matrix set $A = \{ {A^m}\forall m \in M\}$.\\
  \hspace*{0.4in}The node embedding dimension $d$.\\
  \hspace*{0.4in}The number of hybrid metapaths $C=|P|$.\\
  \hspace*{0.4in}The maximum length of hybrid metapath $L$.\\
  \hspace*{0.4in}The number of training rounds $epoch$.\\
  \hspace*{0.02in} {\bf Output:} \\ 
  \hspace*{0.4in}The embedding ${Z_i}$ of node $i$.\\
  \hspace*{0.4in}The weight of each single relationship in different \\ \hspace*{0.4in}hybrid metapaths $\{ w_m^{{\Phi _p}},\forall m \in M,\forall p \in P\}$.\\
  \hspace*{0.4in}The attention of different layers $\{ \beta _{i,l}^{{\Phi _p}},\forall l \in L\}$.\\
  \hspace*{0.4in}The attention of hybrid metapaths $\{ {\beta _{{\Phi _p}}},\forall p \in P\}$.\\
  \begin{algorithmic}[1]
  \For{$e=1,2,...,epoch$} 
    \For{$p=1,2,...,C$}
      \For{$l=1,2,...,L$}
        \State Extract hybrid adjacency matrix $A_{(l)}^{{\Phi _p}}$ \hspace*{0.6in}by HMAE.
        \State Aggregate the information in this hop \hspace*{0.6in}$Z_l^{{\Phi _p}}$ by HLHIA.
      \EndFor
      \State Calculate the attention weights of different hops \hspace*{0.4in}$\beta _{i,l}^{{\Phi _p}}$.
      \State The information of different hops is merged to \hspace*{0.4in}obtain $Z_i^{{\Phi _p}}$ by HLHIA.
    \EndFor
    \State Calculate the attention weights of different hybrid \hspace*{0.2in}metapaths ${\beta _{{\Phi _p}}}$.
    \State The information of different hybrid metapaths is \hspace*{0.2in}merged to obtain ${Z_i}$ by HSAF.
    \State Calculate the loss by equation(12).
    \State Back propagation and update parameters.
    \EndFor
  \Return ${Z_i}$, $\{ w_m^{{\Phi _p}},\forall m \in M,\forall p \in P\}$, $\{ \beta _{i,l}^{{\Phi _p}},\forall l \in L\}$, $\{ {\beta _{i,{\Phi _p}}},\forall p \in P\}$.  
   
  \end{algorithmic}
  \end{algorithm}

\begin{table*}[htb]
\setlength{\abovecaptionskip}{-0.3cm}
\scriptsize
\caption{Datasets of heterogeneous graphs.}
\label{tab2}
\center
\begin{tabular}{@{}ccccccccc@{}}
\toprule
Data               & Relations(A-B)    & Number of A   & Number of B  & Number of A-B & Feature dimension   & Training   & Validation    & Test                \\ \midrule

\multirow{2}{*}{DBLP} & Paper-Author(PA)    & 14328 & 4057 & 19645 & \multirow{2}{*}{334}  & \multirow{2}{*}{800} & \multirow{2}{*}{400} & \multirow{2}{*}{2857} \\ \cmidrule(lr){2-5}
                      & Paper-Conf(PC)     & 14328 & 20   & 14328 &                       &                      &                      &                       \\ \midrule
\multirow{2}{*}{IMDB} & Movie-Actor(MA)    & 4661  & 5841 & 13983 & \multirow{2}{*}{1256} & \multirow{2}{*}{300} & \multirow{2}{*}{300} & \multirow{2}{*}{2339} \\ \cmidrule(lr){2-5}
                      & Movie-Director(MD) & 4661  & 2270 & 4661  &                       &                      &                      &                       \\ \midrule

\multirow{2}{*}{ACM}  & Paper-Author(PA)    & 3025  & 5912 & 9936  & \multirow{2}{*}{1902} & \multirow{2}{*}{600} & \multirow{2}{*}{300} & \multirow{2}{*}{2125} \\ \cmidrule(lr){2-5}
                      & Paper-Subject(PS)  & 3025  & 56   & 3025  &                       &                      &                      &                       \\ \bottomrule
\end{tabular}
\vspace{-2em}
\end{table*}

\section{Experiments}\label{sec:Experiments}

To verify the performance of the model proposed in this paper, we conduct a variety of experiments in this section. First, to investigate the superiority of the proposed model, MHNF is compared with various state-of-the-art baseline methods used in node classification and clustering experiments. Second,  the improvement effect of the extracted hybrid metapaths on the target task is explored. Furthermore, we quantitatively analyze the attention weights of each single relationship in different hybrid metapaths and the attention weights of different hops in the hybrid metapath to verify the model's hybrid metapath extraction ability and the importance of different-hop neighborhood information. In addition, we perform visualization experiments on the node representation obtained by the model to analyze the quality of its spatial distribution. Finally, the hyperparameter study is presented in Section 5.11 to explore the influence of the parameter settings of the networks on the model performance.

\subsection{Datasets.}

The three open-source heterogeneous graphs used in the experiments are commonly used datasets in the field of graph representation learning. The detailed information is shown in Table 2.

\begin{itemize}

	\item \textbf{DBLP\footnote{https://dblp.uni-trier.de}.} It is a citation network dataset that is commonly used in heterogeneous graph representation learning tasks. The nodes in this dataset include paper, author, and conference information. The relationships between the nodes include paper-author and paper-conference. In the experiment, we extracted part of the data in DBLP and divided the papers into four categories: database, artificial intelligence, information extraction, and data mining.
	\item \textbf{IMDB\footnote{https://www.imdb.com}.} It is a movie dataset. The nodes include movie, actor, and director information. The relationships include movie-actor and movie-director. In our experiment, we extracted 4661 movie nodes, 5841 actor nodes, and 2270 director nodes. Nodes are divided into three categories according to the movie plot, namely, action movies, comedies, and dramas. The description of movie plots in the dataset is word embedding obtained by the bag-of-words model, which is used as the initial feature of the movies.
	\item \textbf{ACM\footnote{http://dl.acm.org}.} Like the DBLP dataset, it is also a well-known open-source citation network dataset, but its data scale is smaller. The nodes include paper, author, subject information and the edge includes paper-author and paper-topic relationships. In our experiment, we extracted the papers published in the conferences VLDB, COMM, KDD, SIGMOD, SIGCOMM, and MobiCOMM. We then divided the papers into three categories: database, wireless communication, and data mining. The constructed heterogeneous graph contains 3025 paper nodes, 5912 author nodes, and 56 subject nodes. The initial features of the papers are obtained by embedding the keywords of the paper using the bag-of-words model.
\end{itemize}

\subsection{Baselines.}

In the experiments, the model proposed in this paper is compared with various state-of-the-art methods in the field of graph representation learning, including homogeneous graph neural networks and heterogeneous graph neural networks.

\begin{itemize}
	
	\item \textbf{GCN}\cite{kipf2016semi}: A pioneering work based on a homogeneous GNN model. It extends the original CNN to the graph structure data. Each layer of the model aggregates the one-hop neighbor information of the target node according to a certain strategy (averaging, summation, etc.), which can capture the local structure and node features of the graph.
	
	\item \textbf{GAT}\cite{velivckovic2017graph}: A GNN model of homogenous graphs. It introduces an attention mechanism to calculate the affinity between the target node and its one-hop neighbors. Then, it uses the calculated attention as the weight to aggregate the neighborhood information. This method not only strengthens the aggregation of important neighbor information but also reduces the interference of the noise information from unimportant neighbor nodes on the final node representation. This significantly improves the performance of the model in downstream tasks.
	
	\item \textbf{HAN}\cite{wang2019heterogeneous}: It is a representative work of the latest heterogeneous GNNs. This model extends the GAT model and introduces semantic-level attention to fuse the representation under different metapaths. Therefore, it can better integrate richer semantic information of different metapaths.
	
	\item \textbf{HetGNN}\cite{zhang2019heterogeneous}: It is a state-of-the-art heterogeneous GNN that jointly considers heterogeneous structural (graph) information as well as heterogeneous content information of each node effectively.
	
	\item \textbf{GTN}\cite{yun2019graph}: This model learns soft composite relations for generating useful multihop metapaths and automatically controlling the length. This is similar to the hybrid metapath in this paper. However, our method leverages layer attention to aggregate different hop layer representations and thus does not need to add identity to the relation matrix set as is the case for GTN. More importantly, our hierarchical semantic attention fusion model can efficiently integrate different-hop and different-path neighborhood semantic information. This will be verified in the experiments (Subsections 5.4 and 5.5).
	
	\item \textbf{HGT}\cite{hu2020heterogeneous}: This is also a very recent work. To model heterogeneity, the author designs dependent parameters for different types of nodes and edges to characterize the heterogeneous attention over each edge. This empowers HGT to maintain dedicated representations for different types of nodes and edges. 
	
	\item \textbf{MAGNN}\cite{fu2020magnn}: A latest and state-of-the-art heterogeneous graph embedding method. Intra-metapath information aggregation and inter-metapath information aggregation are achieved successively through the attention neural network. However, we believe that it does not pay attention to the difference between the neighbors of different hops and the correlation between the same hops. Most importantly, the metapath of MAGNN still needs to be specified manually and cannot be learned based on the target task.

	\item \textbf{MHNF}: The method proposed in this article.
	
\end{itemize}
\begin{table*}[t]
  \setlength{\abovecaptionskip}{-0.2cm} 
  \caption{Comparison results of node classification experiment(\%). Bold for "the best", and underline for "the second best".}
  \label{tab3}
  \setlength\tabcolsep{2.2pt}
  \renewcommand{\arraystretch}{1.3}
  \scriptsize
  \center
  \begin{tabular}{@{}ccccccccccc@{}}
  \toprule
  Dataset	& Metrics		& Train(KNN)	
  & HetGNN  & GCN		& GAT  	
  & HGT				& HAN     & MAGNN  & GTN   	
  & \textbf{MHNF}\\ \midrule 

  \multirow{12}{*} {DBLP}
  & \#params(M)	& -  
  & 0.1223		  & 0.0258		& \underline{0.0218}
  & 0.1816 		& 0.0518 		  & 1.0088     	& 0.0300 
  & \textbf{0.0088}  \\ \cmidrule(l){2-11} 
  
  & \#FLOPs(M)	& -
  & 339.8634	  & \textbf{1.0386}		& 132.8290
  & 1349.6838 	& 282.1497 	  & 13226.4812   & 155.4854 
  &  \underline{16.6749}  \\ \cmidrule(l){2-11} 
  
  & \multirow{4}{*}{Macro-F1} & 20\%     
  & $88.24\pm{0.23}$		  & $90.96\pm{0.31}$		& $90.05\pm{0.81}$
  & $89.95\pm{1.40}$				& $91.66\pm{0.30}$ 		  & \underline{$92.02\pm{0.48}$} &$90.46\pm{0.57}$  
  &  \bm{$92.92\pm{0.96}$}       \\ 
  
  &                           & 40\%  
  & $88.64\pm{0.22}$		  & $91.37\pm{0.22}$		& $91.20\pm{0.52}$
  & $90.29\pm{1.31}$				& $91.88\pm{0.17}$ 		  & \underline{$92.17\pm{0.40}$} &$90.69\pm{0.38}$  
  &  \bm{$93.19\pm{0.83}$}       \\
  
  &                           & 60\%
  & $88.66\pm{0.31}$		  & $91.61\pm{0.34}$		& $91.35\pm{0.33}$
  & $90.51\pm{1.31}$				& $92.09\pm{0.19}$ 		  & \underline{$92.20\pm{0.59}$} &$90.99\pm{0.29}$  
  &  \bm{$93.31\pm{0.84}$}       \\
  
  &                           & 80\%        
  & $88.79\pm{0.46}$		  & $91.86\pm{0.70}$		& $91.44\pm{0.59}$
  & $90.89\pm{1.39}$				& $92.10\pm{0.33}$ 		  & \underline{$92.17\pm{0.40}$} &$90.93\pm{0.40}$  
  &  \bm{$93.53\pm{0.87}$}      \\ \cmidrule(l){2-11} 
  
  & \multirow{4}{*}{Micro-F1} & 20\%
  & $89.30\pm{0.21}$		  & $91.84\pm{0.28}$		& $91.70\pm{0.77}$
  & $91.65\pm{1.07}$				& $92.61\pm{0.26}$ 		  & \underline{$92.92\pm{0.42}$} &$91.08\pm{0.52}$  
  &  \bm{$93.83\pm{0.32}$}      \\ 
          
  &                           & 40\%       
  & $89.68\pm{0.20}$		  & $92.18\pm{0.18}$		& $92.15\pm{0.45}$
  & $91.87\pm{0.99}$				& $92.78\pm{0.16}$ 		  & \underline{$93.06\pm{0.37}$} &$91.28\pm{0.37}$  
  &  \bm{$94.01\pm{0.40}$}      \\
    
  &                           & 60\%     
  & $89.75\pm{0.30}$		  & $92.43\pm{0.32}$		& $92.31\pm{0.42}$
  & $92.04\pm{0.99}$				& $93.00\pm{0.18}$ 		  & \underline{$93.10\pm{0.53}$} &$91.56\pm{0.29}$  
  &  \bm{$94.20\pm{0.71}$}      \\
     
  &                           & 80\%    
  & $89.90\pm{0.38}$		  & $92.66\pm{0.68}$		& $92.37\pm{0.48}$
  & $92.42\pm{1.07}$				& $93.07\pm{0.28}$ 		  & \underline{$93.06\pm{0.39}$} &$91.49\pm{0.39}$  
  &  \bm{$94.39\pm{0.42}$}      \\ \midrule 

  \multirow{12}{*} {IMDB}
  & \#params(M)	& -  
  & 0.1223		  & 0.0848		& \underline{0.0808}
  & 0.4756 		& 0.1698 		  & 1.0352     	& 0.0889 
  & \textbf{0.0088}  \\ \cmidrule(l){2-11} 
  
  & \#FLOPs(M)	& -
  & 3002.1252	  & \textbf{0.8949}	& 376.5252
  & 2093.6826 	& 828.7592 	  & 6681.3470   & 107.0805 
  &  \underline{12.4144}  \\ \cmidrule(l){2-11} 

  & \multirow{4}{*}{Macro-F1} & 20\%   
  & $43.77\pm{0.74}$		  & $46.40\pm{0.72}$		& $49.10\pm{0.14}$
  & $51.70\pm{0.60}$				& $54.59\pm{2.03}$ 		  & $56.80\pm{1.72}$     	& \bm{$58.11\pm{1.98}$} 
  &  \underline{$56.87\pm{0.40}$}      \\
  
  &                           	& 40\%       
  & $44.11\pm{0.96}$		  & $46.70\pm{1.16}$		& $50.05\pm{0.35}$
  & $51.57\pm{1.01}$				& $56.65\pm{1.20}$ 		  & $57.44\pm{1.59}$ &\underline{$57.37\pm{1.84}$} 
  &  \bm{$57.58\pm{0.35}$}      \\
  
  &                           & 60\%        
  & $44.44\pm{1.08}$		  & $46.95\pm{0.90}$		& $51.31\pm{0.55}$
  & $51.69\pm{1.26}$				& $58.3\pm{0.95}$ 		 & \underline{$57.56\pm{1.88}$} &$57.31\pm{2.24}$ 
  &  \bm{$57.97\pm{0.53}$}       \\ 
  
  &                           & 80\%       
  & $44.31\pm{1.07}$		  & $47.57\pm{0.72}$		& $51.72\pm{1.16}$
  & $51.47\pm{1.23}$				& $58.38\pm{1.11}$ 		 & \underline{$58.88\pm{2.26}$} &$58.42\pm{2.74}$ 
  &  \bm{$58.92\pm{0.67}$}        \\ \cmidrule(l){2-11} 
  
  & \multirow{4}{*}{Micro-F1} & 20\% 
  & $50.81\pm{0.86}$		  & $52.77\pm{0.68}$		& $54.64\pm{0.25}$
  & $56.54\pm{0.64}$				& $59.16\pm{1.43}$ 		 & \underline{$60.77\pm{2.00}$} &$60.10\pm{2.06}$ 
  &  \bm{$60.83\pm{0.32}$}       \\ 
  
  &                           & 40\%       
  & $51.29\pm{0.99}$		  & $53.03\pm{1.10}$		& $55.69\pm{0.26}$
  & $56.33\pm{1.08}$				& $60.83\pm{0.88}$ 		 & \underline{$61.37\pm{1.35}$} &$60.32\pm{1.75}$ 
  &  \bm{$61.47\pm{0.39}$}       \\ 
  
  &                           & 60\%       
  & $51.46\pm{1.09}$		  & $56.19\pm{0.95}$		& $56.66\pm{0.46}$
  & $56.50\pm{1.05}$				& $62.35\pm{0.86}$ 		 & \underline{$61.77\pm{1.35}$} &$60.33\pm{2.39}$ 
  &  \bm{$61.88\pm{0.41}$}        \\ 
  
  &                           & 80\%      
  & $51.71\pm{0.99}$		  & $53.71\pm{0.79}$		& $57.30\pm{0.75}$
  & $56.32\pm{1.16}$				& $63.44\pm{1.10}$ 		 & \underline{$62.76\pm{2.09}$} &$60.33\pm{2.92}$ 
  &  \bm{$62.85\pm{0.48}$}        \\ \bottomrule 
  \multirow{12}{*} {ACM}
  & \#params(M)	& -  
    & 0.1223		  & 1.2261		& \underline{0.1221}
  & 0.4826 		& 0.2525 		  & 0.6722     	& 0.1303 
  & \textbf{0.0087}  \\ \cmidrule(l){2-11} 
  
   & \#FLOPs(M)	& -
  & 2548.7809	  & \textbf{0.5808	}	& 369.3107
  & 1464.4602 	& 805.2597 	  & 3743.4656   & 75.4057 
  &  \underline{6.4037}  \\ \cmidrule(l){2-11} 
  
  & \multirow{4}{*}{Macro-F1}	& 20\%  
  
  & $86.92\pm{0.20}$		  & $84.83\pm{0.89}$		& $87.87\pm{0.56}$
  & $90.88\pm{0.26}$ 	& $87.94\pm{1.48}$ & $90.79\pm{0.66}$     		& \bm{$92.57\pm{0.13}$} 
  &  \underline{$91.44\pm{0.17}$}  \\ 
  
    &      & 40\%        
  & $87.73\pm{0.45}$		  & $85.30\pm{1.08}$		& $88.44\pm{0.46}$
  & $90.99\pm{0.20}$				& $88.68\pm{1.05}$ 		  & $90.96\pm{0.53}$     	& \bm{$92.53\pm{0.08}$} 
  &  \underline{$91.46\pm{0.18}$}       \\ 
  
  &   		& 60\% 
  & $87.86\pm{0.47}$		  & $85.73\pm{1.06}$		& $89.25\pm{0.37}$
  & $90.99\pm{0.33}$				& $89.15\pm{1.01}$ 		  & $91.03\pm{0.77}$     	& \bm{$92.84\pm{0.04}$} 
  &  \underline{$91.62\pm{0.26}$}       \\ 
  
  &         & 80\%     
  & $88.08\pm{0.65}$		  & $85.91\pm{1.05}$		& $89.55\pm{0.80}$
  & $90.89\pm{0.54}$				& $89.80\pm{0.72}$ 		  & $91.08\pm{0.88}$     	& \bm{$92.65\pm{0.22}$} 
  &  \underline{$91.80\pm{0.31}$}      \\ \cmidrule(l){2-11} 

  & \multirow{4}{*}{Micro-F1} & 20\%       
  & $86.72\pm{0.20}$		  & $84.63\pm{0.90}$		& $87.69\pm{0.58}$
  & $90.83\pm{0.27}$				& $87.67\pm{1.53}$ 		  & $90.63\pm{0.69}$     	& \bm{$92.44\pm{0.15}$} 
  &  \underline{$91.32\pm{0.17}$}       \\  
                                              
  &   		& 40\%        
  & $87.55\pm{0.46}$		  & $85.10\pm{1.09}$		& $88.28\pm{0.51}$
  & $90.78\pm{0.21}$				& $88.42\pm{1.07}$ 		  & $90.78\pm{0.55}$     	& \bm{$92.55\pm{0.09}$} 
  &  \underline{$91.55\pm{0.19}$}        \\
  
  &        	& 60\%       
  & $87.69\pm{0.49}$		  & $85.51\pm{1.06}$		& $89.12\pm{0.37}$
  & $90.84\pm{0.33}$				& $88.92\pm{1.07}$ 		  & $90.86\pm{0.81}$     	& \bm{$92.56\pm{0.03}$} 
  &  \underline{$91.53\pm{0.26}$}        \\ 
  
  &        	& 80\%        
  & $87.94\pm{0.68}$		  & $85.76\pm{1.06}$		& $89.43\pm{0.82}$
  & $90.99\pm{0.53}$				& $89.60\pm{0.74}$ 		  & $90.89\pm{0.91}$     	& \bm{$92.91\pm{0.23}$} 
  &  \underline{$91.68\pm{0.29}$}        \\ \midrule 
  
  \vspace{-0.8cm}
  \end{tabular}
  \end{table*}
  \begin{table*}[t]
    \setlength{\abovecaptionskip}{-0.2cm} 
    \caption{Comparison results of node clustering experiment(\%).Bold for "the best", and underline for "the second best".}
    \label{tab14}
    \setlength\tabcolsep{2.2pt}
    \renewcommand{\arraystretch}{1.3}
    \scriptsize
    \center
    \begin{tabular}{ccccccccccc}
    \hline
    Datasets              & Metrics 	 	& HetGNN 	& GCN    	& GAT 	   & HGT   & HAN   	& MAGNN & GTN &  \textbf{MHNF}            \\ \hline 

    \multirow{6}{*}{DBLP} 
    
    & \#params(M)
    & 0.1223		  & 0.0258		& \underline{0.0218}
    & 0.1816   		& 0.0518   	& 1.0088  & 0.0300
    & \textbf{0.0088}  \\ \cmidrule(l){2-10} 
    
    & \#FLOPs(M)	
    & 339.8634	  & \textbf{1.0386}		& 132.8290
    & 1349.6838  & 282.1497   & 13226.4812 & 155.4854
    &  \underline{16.6749}  \\ \cmidrule(l){2-10} 

    & NMI    
    & $66.61\pm{0.33}$		  & $71.55\pm{0.96}$		& $74.22\pm{0.48}$
    & $72.26\pm{3.31}$  		& $75.49\pm{1.02}$		&\underline{$77.01\pm{1.28}$} & $21.12\pm{1.08}$
    &  \bm{$77.27\pm{0.82}$}   \\ \cmidrule(l){2-10}
    
    & ARI   
    & $73.48\pm{0.25}$		  & $76.31\pm{0.81}$		& $79.43\pm{0.15}$
    & $79.62\pm{2.85}$ 		& $81.32\pm{1.22}$		 &\underline{$81.39\pm{0.93}$} & $10.83\pm{0.13}$
    &  \bm{$82.10\pm{0.67}$}    \\ \hline 

    \multirow{6}{*}{IMDB} 
    
    & \#params(M)
    & 0.1223		  & 0.0848		& \underline{0.0808}
    & 0.4756     		&  0.1698 		  	&  1.0352  & 0.0889 		
    & \textbf{0.0088}  \\ \cmidrule(l){2-10} 
    
    &\#FLOPs(M)	
    & 3002.1252	  & \textbf{0.8949}		& 376.5252
    & 2093.6826   	&  828.7592 	  &  6681.3470  & 107.0805 	
    &  \underline{12.4144}  \\ \cmidrule(l){2-10} 

    &NMI   
    & $7.78\pm{1.03}$		  & $9.59\pm{0.67}$		& $10.02\pm{0.38}$
    & $13.08\pm{0.29}$ 		& $13.08\pm{1.11}$		  &$15.59\pm{2.01}$  & \bm{$18.69\pm{0.33}$} 		
    &  \underline{$15.63\pm{0.02}$}     \\ \cmidrule(l){2-10}
    
    & ARI
    & $7.44\pm{0.91}$		  & $6.59\pm{0.82}$		& $8.69\pm{0.81}$
    & $14.00\pm{1.48}$ 		& $10.94\pm{1.54}$		 &$13.36\pm{2.27}$  		& \bm{$18.68\pm{0.56}$} 	
    &  \underline{$14.56\pm{0.01}$}    \\ \hline
    
    \multirow{6}{*}{ACM}  
    
    & \#params(M)	
      & 0.1223		  & 1.2261		& \underline{0.1221}
    & 0.4826    		&  0.2525 		  &  0.6722    & 0.1303 		
    & \textbf{0.0087}  \\ \cmidrule(l){2-10} 
    
    & \#FLOPs(M)	
    & 2548.7809	  & \textbf{0.5808	}	& 369.3107
    & 1464.4602   	& 805.2597 	  & 3743.4656  & 75.4057 	
    &  \underline{6.4037}  \\ \cmidrule(l){2-10} 
    
    & NMI   
    
    & $48.77\pm{0.74}$		  & $58.78\pm{0.87}$		& $63.19\pm{1.99}$
    & $71.26\pm{0.53}$ 		& $66.49\pm{2.04}$			&$72.03\pm{2.35}$  	& \bm{$74.92\pm{1.81}}$ 	
    &  \underline{$73.02\pm{0.01}$}  \\  \cmidrule(l){2-10}  
    
     & ARI  
    & $44.11\pm{0.92}$		  				& $62.65\pm{1.15}$			& $67.75\pm{1.98}$
    & $76.62\pm{0.59}$ & $70.56\pm{2.24}$			 &$76.56\pm{2.53}$   	& \bm{$79.80\pm{1.97}}$ 		
    &  \underline{$77.99\pm{0.02}$}   \\ \hline 
    
    \vspace{-0.8cm}
    \end{tabular}
    \end{table*}
\subsection{Implementation Details.}

The parameters of the MHNF model are initially randomized, and the model is optimized by Adam\cite{kingma2014adam}. The learning rate is set to 0.05, weight decay is set to 0.001, the number of hybrid metapaths C is searched in the range of \{1,2,3,4,5\} and in most cases the optimal value is 2. For the ACM and DBLP datasets, the optimal maximum path length (L) is selected as 3, and for the IMDB dataset, the optimal value is selected as 4. For methods such as GCN and HAN, this paper takes the initially obtained node attribute vector as input and adopts the averaging strategy to aggregate neighborhood information. For methods that require manual specification of metapaths (such as HAN, MAGNN, etc.), we use the preset metapaths shown in Table 5. For the selection of hyperparameters for all baseline methods, we refer to the settings in the original papers and leverage the grid search method to determine the optimal parameters. For all baselines, we set the dropout rate to 0.5 and  we employ the Adam optimizer with the learning rate set to 0.005 and the weight decay (L2 penalty) set to 0.001. In particular, for the homogenous graph representation learning methods (GCN, GAT), the experimental results are the best results among all specified metapaths. Unless otherwise stated, other hyperparameters are the default parameters of the original settings of the baselines. The embedding dimension of nodes is set to 64. To ensure the stability of the results, all the reported results are the average results of 10 experiments.

All experiments were carried out on a machine that has an Nvidia RTX 3090 GPU, 128 GB of memory, AMD CPU 3900X @3.80 GHz 24-core, 4 TB disk space, running Ubuntu GNU/Linux. The HetGNN used in this paper is available at the OpenHGNN library \footnote{\url{https://github.com/BUPT-GAMMA/OpenHGNN}}, MAGNN and GTN are available in the original papers, and the other baselines are available at the DGL library \footnote{\url{https://github.com/dmlc/dgl}}. The model proposed in this paper is implemented on PyTorch, and our code is publicly available at \url{https://github.com/PHD-lanyu/MHNF}.

\subsection{Analysis of Node Classification Results.}

In the node classification task experiment, the low-dimensional vectors of nodes obtained by each method are fed into KNN, and the node labels (such as the category of papers) are used for the classification prediction task. First, different proportions of nodes are used to train KNN, and the rest are used as test data.
This paper uses both Micro-F1 and Macro-F1 metrics for evaluation, which can verify the classification effect and quality in a more balanced and comprehensive manner. The obtained node classification experiment results are shown in Table 3\footnote{It is noted that the classification metrics on the DBLP dataset of GTN in our manuscript are lower than those of the original paper, the reason is that we are conducting the experiments in a two-stage setup but GTN is an end-end manner.}.

It can be seen from Table 3 that the method proposed in this paper (MHNF) consistently achieves the best or competitive performance on both Macro-F1 and Micro-F1 metrics under different training ratios of KNN. The performance gaps between the MHNF, MAGNN and GTN are not significant (even smaller than the standard deviation in some cases), this has also been verified by significance test (unpaired t-test) experiments (A similar phenomenon also occurs in Table 4.). Due to space limitations, we do not show the details of the significance test. Howerver, the MHNF method outperforms or achieves comparative performance against GTN and MAGNN with 1/10 $\sim$ 1/100 parameters and computational budget. We attribute it to both the ability of the HMAE model (Section 4.1) to learn the hybrid metapath instead of manually specifying the metapaths and hierarchical semantic attention fusion (Sections 4.2 and 4.3) over multihop neighbors and different hybrid metapaths. These features are not available in the baseline methods. Although GTN performs slightly better than MHNF on ACM datasets, this comes at the cost of more than 10 times more parameters and computational budget.
Compared with the other baseline methods, MHNF brings significant improvements against both evaluation metrics with much fewer training parameters and FLOPs. 

It can be clearly seen from the results that the homogeneous GNNs (GCN and GAT) have relatively poor performance compared with the heterogeneous GNNs. This reflects the ability of heterogeneous GNNs to integrate more diverse graph structures and node features. GAT outperforms GCN in most cases, which shows that the attention mechanism between nodes has a positive effect on classification tasks. GTN and MAGNN achieved the best results among all baseline methods, which shows the superiority of hybrid metapaths and the importance of inner-metapath node information. As for the original paper of HetGNN, it has great performance in aggregating heterogeneous content information (e.g., text, images) of nodes, but in our experiments, it does not perform as expected in aggregating heterogeneous interactions between nodes. 

\subsection{Analysis of Node Clustering Results.}

To comprehensively evaluate and compare the spatial distribution of the low-dimensional vectors obtained by each model, we fed the low-dimensional node embeddings into the k-means clustering model to conduct a node clustering experiment. In the experiment, the number of clusters was set to be the same as the node category of each dataset. We chose NMI and ARI which are the commonly used standard evaluation protocols in clustering tasks as evaluation metrics. 

The results of node clustering experiments are shown in Table 4. The MHNF model consistently outperforms the baseline methods by a large margin, except for being slightly worse than GTN on the IMDB and ACM datasets. It is worth noting that GTN achieves its performance using more than 10 times the number of parameters and FLOPs than MHNF. The performance of GTN on the DBLP dataset shows a catastrophic drop. We suspect the possible reason is it's overfocusing on the node classification task (the target task of the training phase), which affects its performance on node clustering. Compared with GTN, MHNF consistently achieves competitive performance on node clustering and node classification tasks. This indicates that the node representation learned by MHNF has better generality for different downstream tasks. At the same time, similar to the results in the node classification experiment, the results of heterogeneous GNNs (such as HAN, GTN, HGT, MAGNN, and MHNF.) are very comparable to homogeneous GNNs (GCN and GAT), further verifying the powerful representation ability of heterogeneous GNNs.

\subsection{Scalability and Computational Efficiency Analysis.}

From the results reported in  Subsections 5.4 and 5.5, we can conclude that the method proposed in this paper achieves the best or most competitive performance. In this section, we will compare and analyze the scalability and computational efficiency of the different methods by combining the node classification and node clustering results. 

As seen in Table 3 and Table 4, MHNF achieves the best performance with the lowest number of parameters and computational budget (approximately 1/100 of MAGNN and 1/10 of GTN). This shows great efficiency and scalability when dealing with datasets of different sizes. GTN and GCN have the lowest number of parameters and FLOPs among all the baseline methods, but the performance of GCN lags far behind that of GTN because it ignores the type of node features and the relationship between nodes. Although MAGNN and HGT achieve competitive performances, the number of parameters and FLOPs of these methods are 10 $\sim$ 100 times higher than those of other methods. This will greatly limit their application in scenarios involving large-scale datasets or stacking multiple layers is required to obtain better performance.

\subsection{The performance with different training/test ratios.}

To show how the performance varies as the training/test ratio varies, we randomly divide the datasets according to different training/testing ratios. Specifically, we first divide a dataset into a training set and a non-training set according to the label distribution of the nodes in a certain ratio (e.g., 2:8) and then use 20\% of the nontraining set as the validation set and the rest as the test set. When using the test set to verify the model performance, we fix the training ratio of KNN to be 40\%.

\begin{figure}[htp]
  \centering
  \includegraphics[width=3.5in,trim=0 40 0 20]{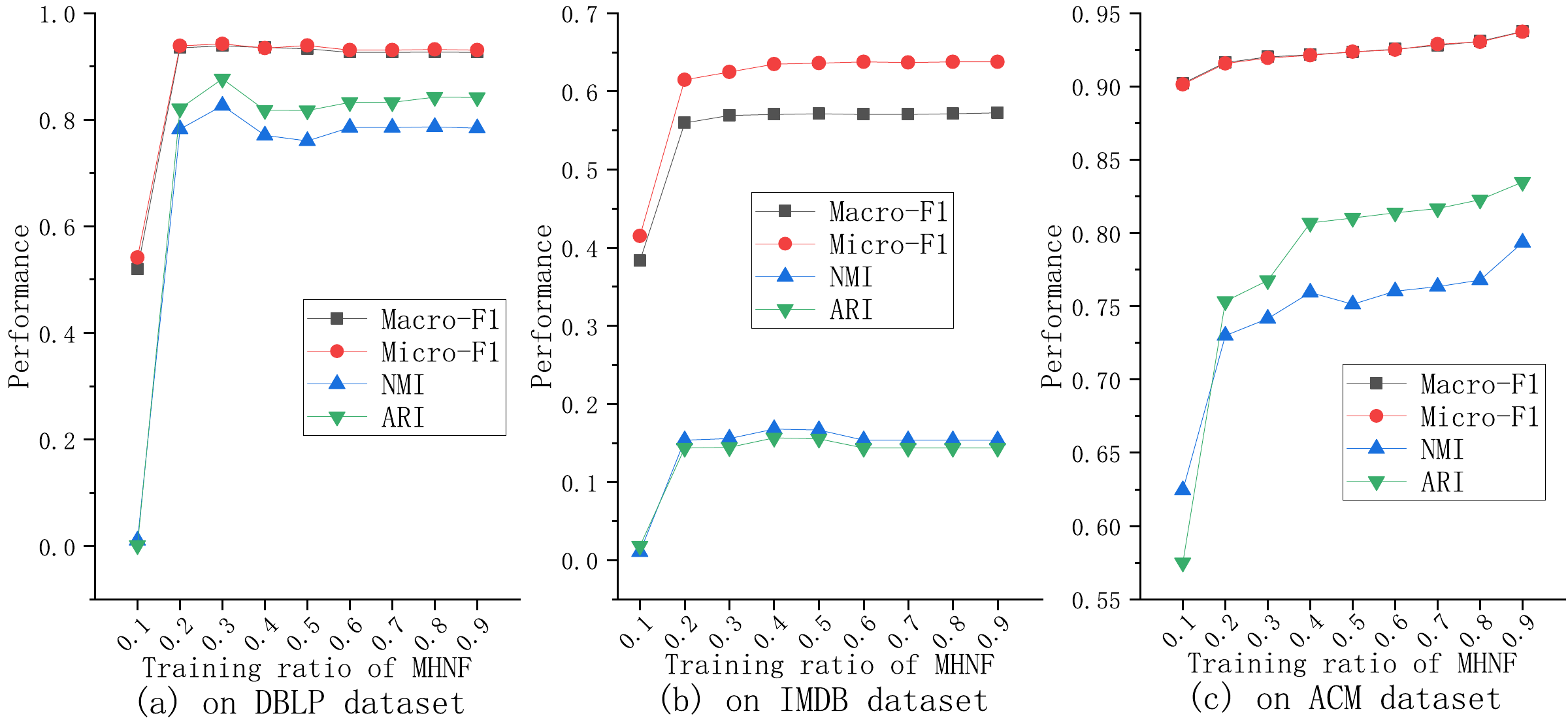}
  \caption{The performance of MHNF with different training/test ratios (the training ratio of KNN is fixed as 40\%).}
  \label{train_ratio}
  \vspace{-0.4cm}
  \end{figure}

The experimental results on the three datasets are shown in Fig.\ref{train_ratio}.  From the figure, it can be seen that the classification and clustering performance of MHNF are poor when the training data ratio is 0.1, indicating that the model is not able to capture the relational patterns of the graph well with too few nodes. When the training ratio reaches 0.2, the performance of the model increases rapidly. When the training ratio continues to increase, the performance of the model does not improve significantly, which indicates that 20\% training nodes are sufficient for the model to mine the relational patterns of the graph.

\subsection{Ablation and Effectiveness Analyses.}

To investigate the ability of the HMAE model (Section 4.1) to extract important metapaths and the improved effect of the extracted hybrid metapaths on the target tasks, the following twofold experiments were carried out.

First, we recorded the fusion weights learned by the autonomous extraction model of the hybrid heterogeneous adjacency. We multiplied the weight of the path and the intrapath level attention and set it to be the weight of the path and selected the path with the top 3 weights, as shown in Table 5. It can be seen that our method can learn the metapath manually specified by domain knowledge. At the same time, the model can also learn metapath information that is not specified. For example, the APAPA path in DBLP can mine high-order cooperative relations, and authors with high-order cooperative relations are most likely to belong to the same research field and this plays an important role in improving node classification and clustering.

\begin{table}[htb]
\setlength{\abovecaptionskip}{-0.2cm}
\caption{The metapath (top3) learned by HMAE on different datasets and the corresponding preset metapath.}
\label{tab4}
\center
\begin{tabular}{@{}ccc@{}}
\toprule
Dataset & Preset Metapath & Metapath learnt by HMAE \\ \midrule
DBLP    & APCPA, APA      & APCPA, APCPC, APAPA     \\ \midrule
IMDB    & MAM, MDM        & MAM, MDM, AMA           \\ \midrule
ACM     & PAP, PSP        & APA, PAP, PSP           \\ \bottomrule
\end{tabular}
\vspace{-0.1cm}
\end{table}

Second, the model variant experiments were carried out on the three datasets. Due to space limitations, we show the results on the ACM and IMDB datasets only. The results for the DBLP dataset show exactly the same trend. Specifically, we manually specify different metapaths to replace MHNF's hybrid metapath extraction model and obtain vanilla versions of the MHNF. For example, on the ACM dataset, PAP and PSP mean specifying the path of PAP and PSP, respectively, and PAP+PSP means that both metapaths are considered. Hybrid-2 represents a hybrid metapath with a maximum length of 2 extracted by the HMAE model.
 
\begin{figure}[t]
  \centering
  \includegraphics[width=3.5in,trim=0 40 0 20]{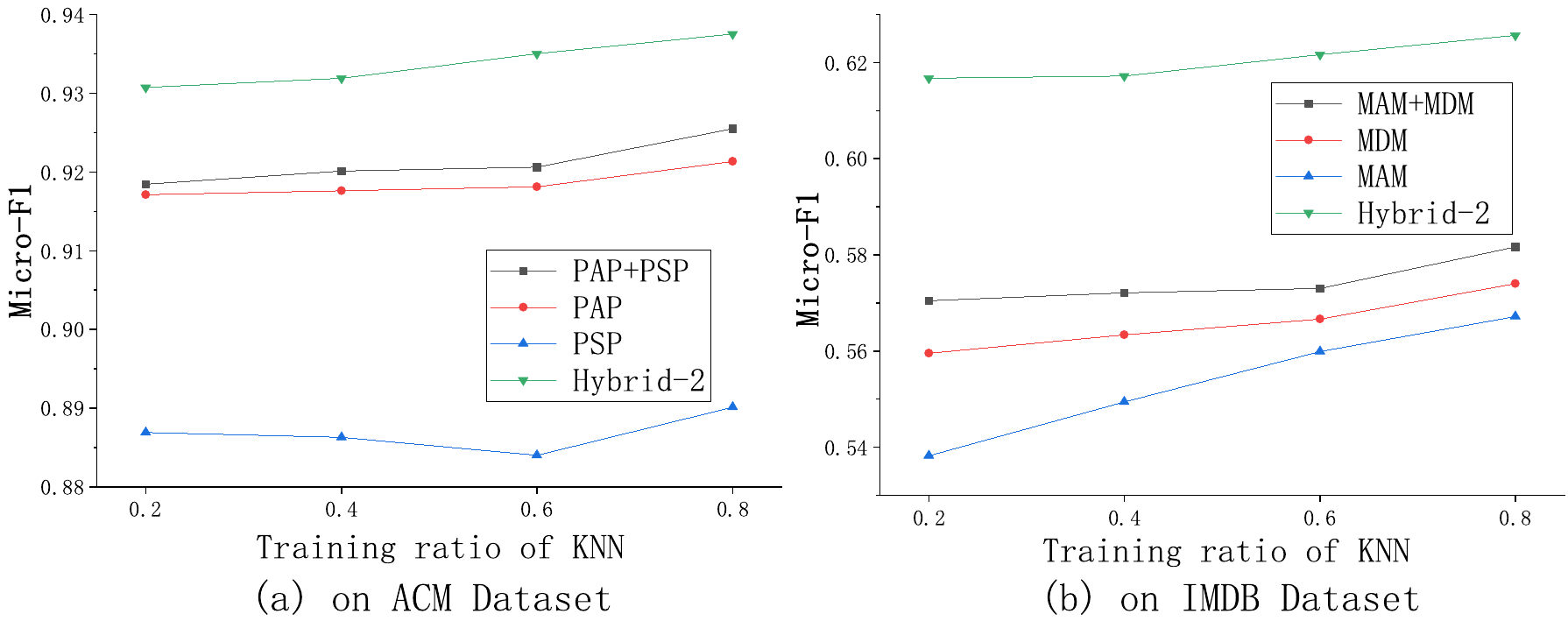}
  \caption{Comparison results of the classification performance of MHNF and its vanilla versions on the ACM and IMDB datasets.}
  \label{figure6}
  \vspace{-0.3cm}
  \end{figure}

The curves of the experimental results are shown in Fig.\ref{figure6}. On both datasets, the HMAE model outperforms the vanilla versions by a large margin. This shows that the HMAE model can autonomously extract valuable hybrid metapaths and it can improve the target task performance. At the same time, on both datasets, the dual-metapath versions (PAP+PSP and MAM+MDM) outperform the corresponding single-path version, which shows that the HLHIA (Section 4.2) and HSAF (Section 4.3) models can efficiently integrate the semantic information of different metapaths and mine deeper and more valuable information for the target task.

\subsection{Analysis of Attention Mechanism.}

\begin{figure}[htp]
  \centering
  \includegraphics[width=3.5in,trim=0 40 0 20]{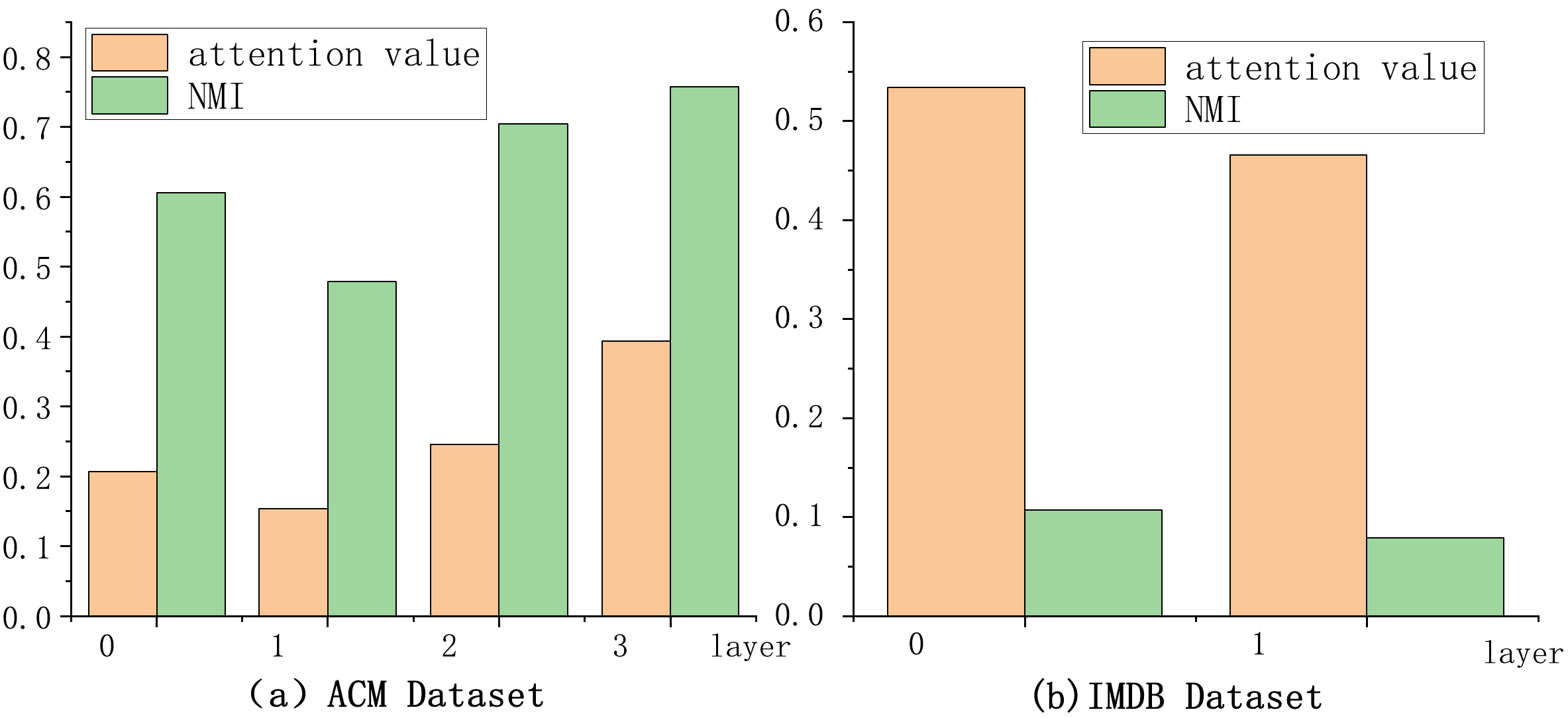}
  \caption{The clustering performance of single-hop information and the attention weight of the corresponding hop on the ACM and IMDB datasets.}
  \label{figure7}
    \vspace{-0.3cm}
  \end{figure}

A notable feature of our method is the introduction of information fusion of inner-metapath nodes (see \textit{Definition3.6}). Therefore, this section analyzes the attention weights of different hops. Fig.\ref{figure7} shows the clustering performance of single-hop information and the attention weight of the corresponding hop. Due to space limitations, we show the results on the ACM and IMDB datasets only.  The results on the DBLP dataset show exactly the same trend. It can be clearly seen that there is a strong positive correlation between the two results. This demonstrates that the model can correctly give higher attention to the important information of the corresponding hops that greatly improve the final performance. At the same time, the higher attention of the middle hops also proves the importance of the inner-metapath nodes, which is consistent with the original assumption that inner-metapath nodes play a pivotal role in precisely mining the complex relationships between the starting node and destination node.
  
\begin{figure*}[t]
  \centering
  \includegraphics[width=6.5in,trim=0 20 0 10]{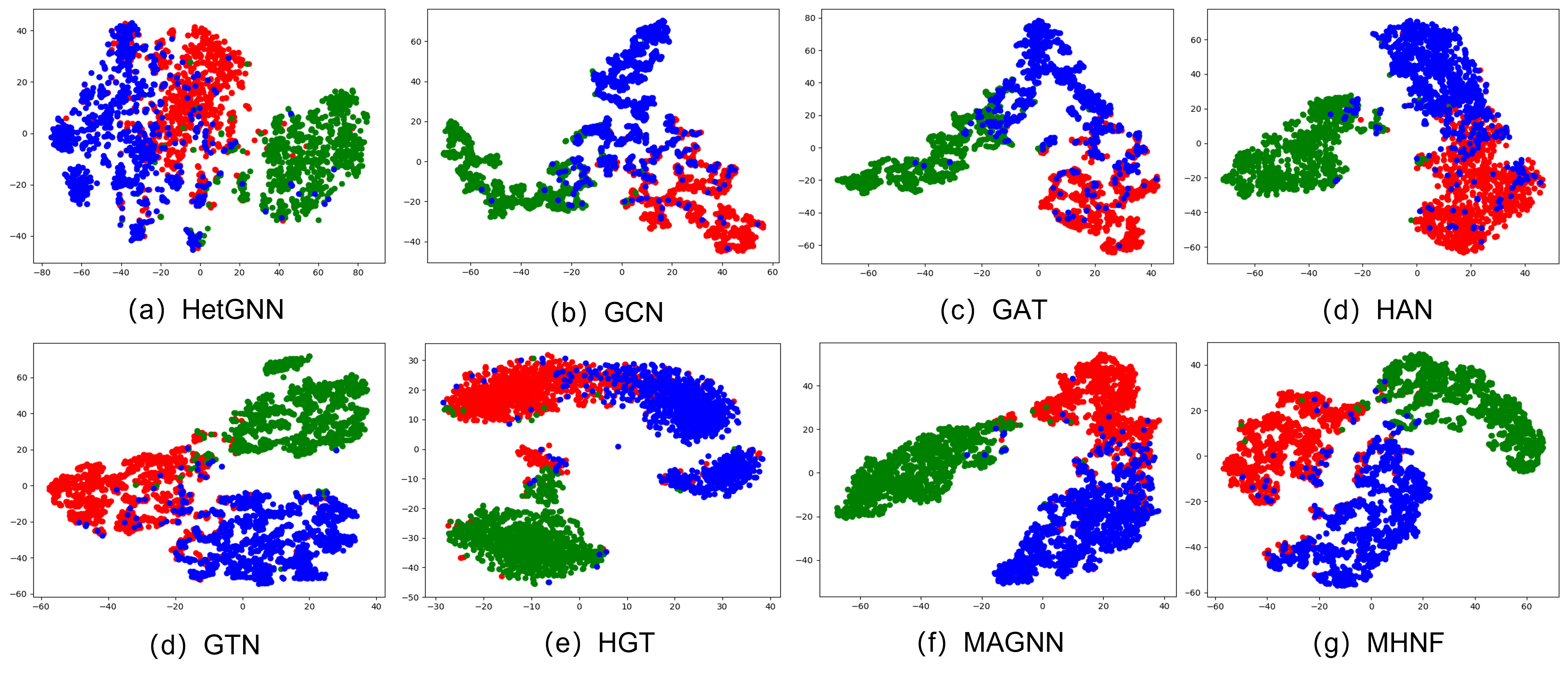}
  \caption{The space distribution diagram of the node embeddings learned by different models on the ACM dataset. Different colors represent different research areas of the papers.}
  \label{figure8}
  \end{figure*}
  
\subsection{Visualization Experiment.}

\begin{figure*}[h]
  \centering
  \includegraphics[width=6.5in,trim=0 40 0 20]{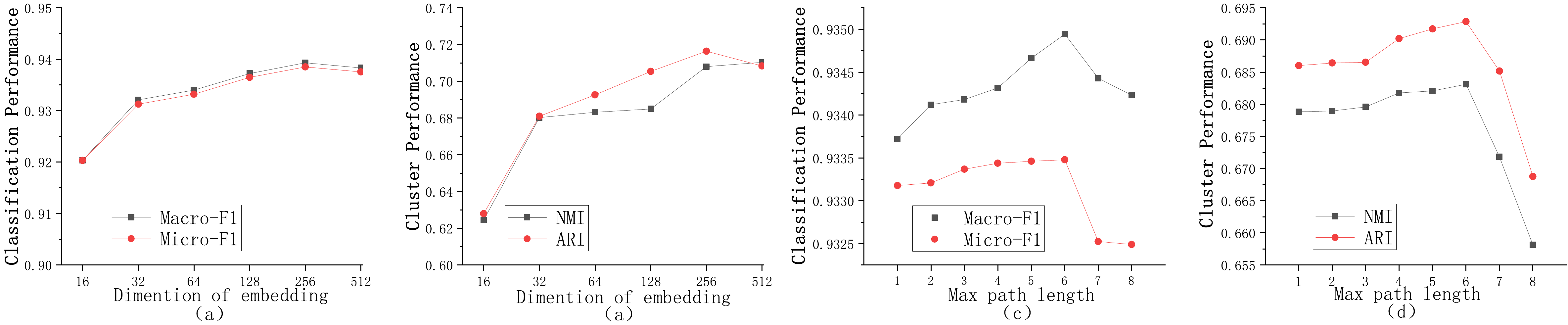}
  \caption{ A schematic diagram of the influence of changes in various hyperparameters on the performance of different tasks. Due to space limitations, we show the results on the ACM dataset only, and the results on the IMDB and DBLP datasets show the same trend. (a)  the impact of different node embedding dimensions on the classification task. (b) the impact of different node embedding dimensions on the clustering task. (c) the impact of the maximum length of the hybrid metapath on the classification task. (d) the impact of the maximum length of the hybrid metapath on the clustering task.}
  \label{figure9}
  \vspace{-0.3cm}
  \end{figure*}  

In this subsection, we conduct visual experiments to intuitively show the distribution of the node embedding learned by the models and the ability to support downstream tasks. We first use t-sne\cite{van2008visualizing} to compress the node embeddings into a two-dimensional space and then visually display it. This is a common evaluation scheme for graph representation tasks. Fig.\ref{figure8} shows the distribution of nodes learned by different models on the ACM dataset. The different colors represent the categories (the fields of the papers in the ACM dataset) of nodes. It can be seen intuitively from Fig.\ref{figure8} that the spatial distribution of the embeddings learned by all methods can reflect the correlation with the node category.

However, the boundaries between the embeddings of different categories obtained by HetGNN, HGT, and GCN are not obvious, and there are a large number of intersecting nodes between each category. Compared with them, GAT and GTN achieve more satisfactory performance. For GAT, the red and green node clusters are clearly separated, but the blue node is mixed with the other two colors. A similar situation occurs with MAGNN. Compared with the above methods, HAN and GTN achieve the best performance among baseline methods. Taking HAN as an example, the green node and the other two colors have a larger distribution interval, but unfortunately, there are some intersecting nodes between the blue and red clusters. 

The embedding distribution obtained by the MHNF method is obviously the best one. Not only is the spatial distribution of nodes highly correlated with category labels, but nodes of different categories rarely have intersecting phenomena. More importantly, the boundaries between all the different node clusters are also obvious. This shows that the model has excellent discrimination for different types of nodes, which will be very conducive to the generalization of downstream classification tasks and clustering tasks. The above results are sufficient to show that the low-dimensional node embeddings learned by the MHNF model will provide strong support for downstream tasks.

\subsection{Hyper-parameter Studies.} 

In this section, we explore the sensitivity of the model by setting different hyperparameters and observing the changes in node classification and node clustering performance. Due to space limitations, we show the results on the ACM datasets only. The results for the IMDB and DBLP datasets show the same trend.
\begin{itemize}
	\item \textbf{Dimension of the final embedding.}

First, to explore the effect of embedding dimensions on task performance, we set dimensions in the range \{16, 32, 64, 128, 256, 512\}. The classification and clustering performance curves are shown in Fig.\ref{figure9}(a) and Fig.\ref{figure9}(b), respectively. Initially, as the dimension increases, classification and clustering performance can be improved, which shows that higher-dimensional representations can cover richer semantic information. However, when the dimensionality is increased to 512, the classification and clustering performances decrease slightly, which shows that representation with too high-dimensional information introduces some noise and this is not conducive to the improvement of the performance.

	\item \textbf{Max path length.} 

Second, to explore the influence of the maximum length of the hybrid metapath on the performance of downstream tasks, we varied the maximum length from 1 to 8. The resulting classification and clustering performance curves are shown in Fig.\ref{figure9}(c) and Fig.\ref{figure9}(d), respectively. Similar to the node dimension experiment, initially, as the maximum path length increases, the classification, and clustering performance improves, which shows that a wider range of nodes can cover richer semantic information. However, when the maximum length is increased to more than 6, the classification and clustering performance obviously decrease, which shows that a path that is too long introduces some noise and this is not conducive to the improvement of the performance. This observation is consistent with the conclusions drawn from previous studies\cite {ji2020dual, wang2019neural, he2020lightgcn}.	
\end{itemize}

\section{Conclusion}

This paper solves a series of key problems arising in network representation learning and GNNs by proposing the MHNF model. First, the hybrid metapath autonomous extraction model is built, which can autonomously learn valuable hybrid metapaths, to achieve efficient extraction of multihops neighbors with different hops. Second, the hop-level heterogeneous information aggregation model within one hybrid metapath realizes the aggregation of information from neighbors of different hops within one path. It is worth noting that the existing methods cannot efficiently aggregate multihop neighborhood information. Finally, the hierarchical semantic attention fusion model realizes the fusion of information within a path as well as different paths. This solves the problem of information aggregation from inner-metapath nodes which the existing metapath-based methods can not deal with. The results of node classification and node clustering experiments show that MHNF achieves the best or competitive performance against state-of-the-art baselines with only 1/10 $\sim$ 1/100 parameters and computational budget. Attention analysis experiments and model variant experiments verify the model's ability to learn hybrid metapaths and to integrate information from neighbors of different hops autonomously and efficiently. The results also verify the importance of inner-metapath nodes, which is in line with the original insights of this paper.

Of course, the method proposed in this paper also has room for optimization. First, in the present paper, the method is implemented on a static graph. However, many realistic graphs are dynamic: the nodes, edges, and features in the graph can change over time. For example, in social networks, people can establish new social relationships, delete old relationships, and update their profiles, such as hobbies and occupations. New users can join one community, and existing users can leave. How to model the evolutionary characteristics of dynamic graphs and support the incremental tuning of model parameters is still largely unresolved. In addition, the study of network representation learning in hyperbolic space is also a very promising direction and this will be the goal of our future research.


%



\ifCLASSOPTIONcompsoc
  \section*{Acknowledgments}
\else
  \section*{Acknowledgment}
\fi

This work is supported by National Key R\&D Program of China (2020YFE0201500), the Fundamental Research Funds for the Central Universities (HIT.NSRIF.201714), Weihai Science and Technology Development Program (2016DXGJMS15), and the Key Research and Development Program in Shandong Province (2017GGX90103).

\ifCLASSOPTIONcaptionsoff
  \newpage
\fi

\bibliography{bib/bib}  

\begin{IEEEbiography}[{\includegraphics[width=1in,height=1.25in,clip,keepaspectratio]{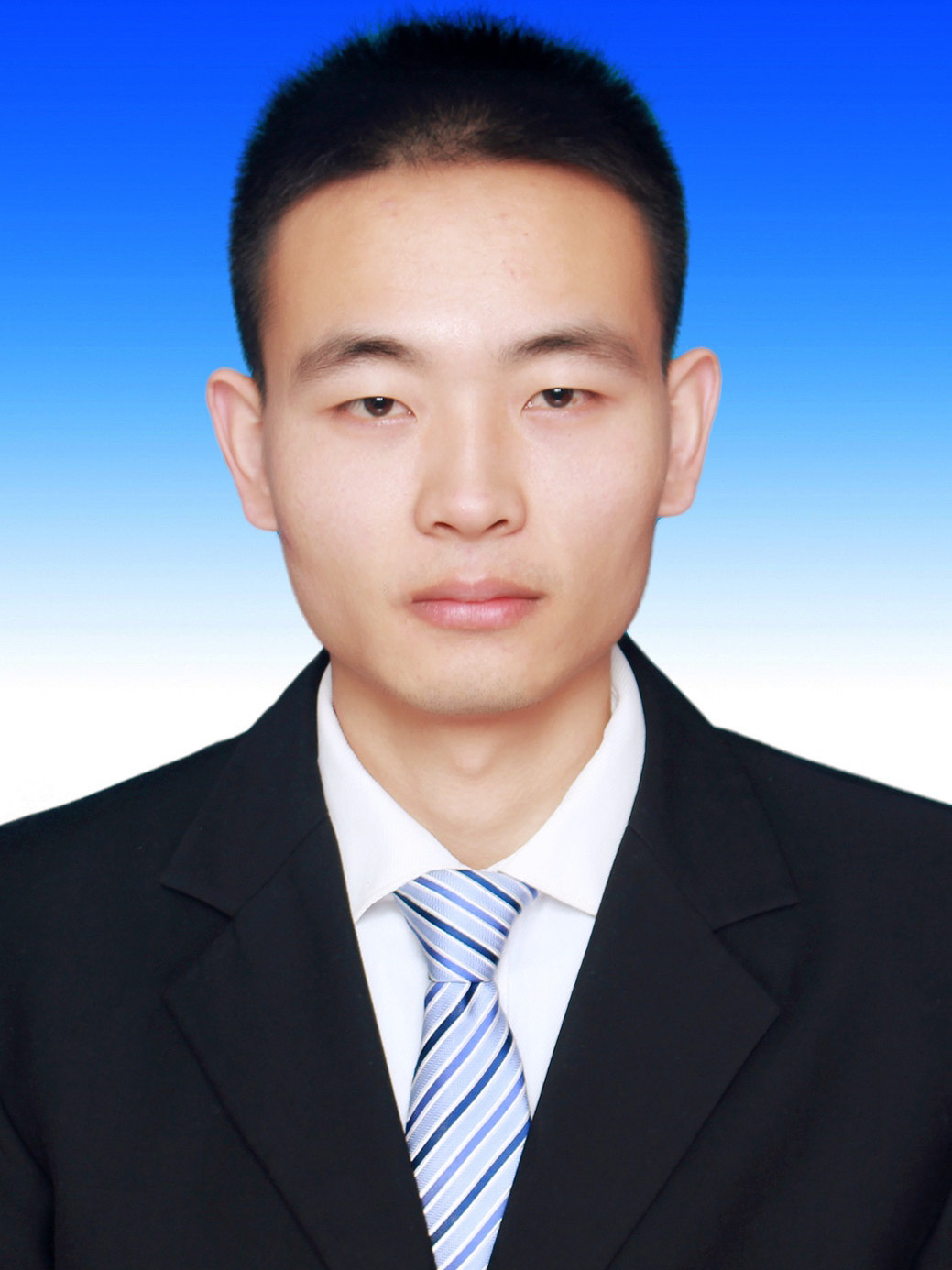}}]{Yundong Sun}
  is currently working toward the Ph.D degree in School of Astronautics at Harbin Institute of Technology. His research interests include machine learning, graph representation learning, recommender system and network embedding.
\end{IEEEbiography}
\begin{IEEEbiography}[{\includegraphics[width=1in,height=1.25in,clip,keepaspectratio]{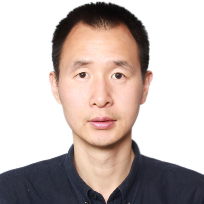}}]{Dongjie Zhu}
  received the Ph.D degree in computer architecture from the Harbin Institute of Technology, in 2015. He is an associate professor in School of Computer Science and Technology at Harbin Institute of Technology, Wehai. His research interests include parallel storage systems, social computing, machine learning.
\end{IEEEbiography}

\begin{IEEEbiography}[{\includegraphics[width=1in,height=1.25in,clip,keepaspectratio]{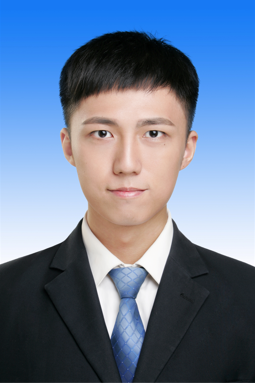}}]{Haiwen Du}
  is currently working toward the Ph.D degree in School of Astronautics at Harbin Institute of Technology. His research interests include storage system architecture and massive data management.
\end{IEEEbiography}

\begin{IEEEbiography}[{\includegraphics[width=1in,height=1.25in,clip,keepaspectratio]{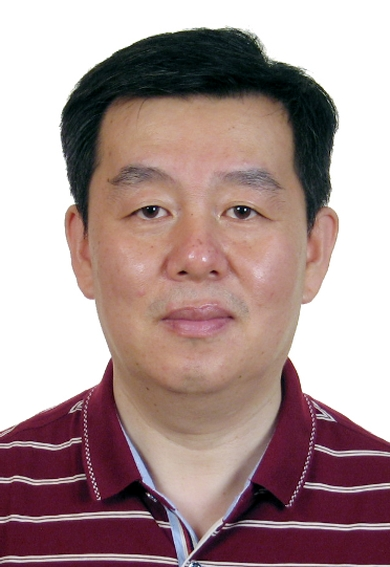}}]{Zhaoshuo Tian}
  is a professor in School of Astronautics at Harbin Institute of Technology. His research interests include laser technology and marine laser detection technology. He is a member of IEEE.

\end{IEEEbiography}
\end{document}